\documentclass{article} 
\usepackage{iclr2024_conference,times}

\usepackage{amsmath,amsfonts,bm}

\def\eqref#1{equation~\ref{#1}}

\def\1{\bm{1}}

\DeclareMathAlphabet{\mathsfit}{\encodingdefault}{\sfdefault}{m}{sl}
\SetMathAlphabet{\mathsfit}{bold}{\encodingdefault}{\sfdefault}{bx}{n}

\usepackage{hyperref}
\usepackage{url}

\usepackage{algorithm}
\usepackage[noend]{algpseudocode}

\usepackage{graphicx}
\usepackage{subcaption}
\usepackage{float}
\usepackage{diagbox}
\usepackage{multirow}
\usepackage{booktabs}

\hypersetup{
colorlinks, linkcolor=red, citecolor=cyan
}

\title{Has your Pre-trained Model Improved? \\A Multi-head Posterior Based Approach}

\author{Prince Aboagye\thanks{These authors contributed equally to this work.  Correspondence email: \texttt{wzhan@visa.com}} \and
  Yan Zheng\footnotemark[1] \footnotemark[2]  \and
  Junpeng Wang\footnotemark[1] \footnotemark[2] \and
  Uday Singh Saini\footnotemark[1] \footnotemark[2]  \and
  Xin Dai\thanks{Visa Research, Palo Alto, CA.}  \and
  Michael Yeh\footnotemark[2] \and
  Yujie Fan\footnotemark[2]  \and
  Zhongfang Zhuang\footnotemark[2] \and
  Shubham Jain\footnotemark[2] \and
  Liang Wang\footnotemark[2] 
  \and Wei Zhang\footnotemark[1] \footnotemark[2] 
}

\iclrfinalcopy 
\begin{document}

\maketitle


\begin{abstract}
The emergence of pre-trained models has significantly impacted Natural Language Processing (NLP) and Computer Vision to relational datasets. Traditionally, these models are assessed through fine-tuned downstream tasks. However, this raises the question of how to evaluate these models more efficiently and effectively. In this study, we explore a novel approach where we leverage the meta-features associated with each entity as a source of worldly knowledge and employ entity representations from the models. We propose using the consistency between these representations and the meta-features as a metric for evaluating pre-trained models. Our method’s effectiveness is demonstrated across various domains, including models with relational datasets, large language models, and image models.
\end{abstract}


\section{Introduction}
The rise in user-generated content has led to the widespread adoption of pre-training large models across numerous machine-learning fields. This trend is particularly noticeable in Natural Language Processing (NLP) with GPT (Generative Pretrained Transformer) models ~\citep{brown2020language} and vision-language domains with models like CLIP ~\citep{radford2021learning}. Usually, the performance of these models is assessed through various downstream tasks such as Machine Translation \citep{wang-etal-2023-document-level}, Factuality \citep{chen-etal-2019-evaluating}, Question answering \citep{liang2023holistic}, Multilingual tasks \citep{bang-etal-2023-multitask, ahuja2023mega}, which can be relatively expensive if a large number of tasks are considered necessary for a robust evaluation. Is there an alternative approach that offers both efficiency and simplicity for evaluating these models?

Entity representations, also known as embeddings generated from these models, can be utilized directly or indirectly by downstream tasks and fine-tuned as needed. The associated meta-features with these embeddings can be considered as the model's foundational knowledge of the world it's learning from. This could be the class category for image data or semantic and syntactic information for words. Despite having the same meta-features, embeddings differ across models. Therefore, the degree of consistency between the embeddings and meta-features can serve as a performance metric for model evaluation.

Embedding spaces are complex and challenging to interpret or explain. Despite extensive efforts to decipher it, its intricacies go beyond mere linear interpretability, as some research suggests.  
In this research, we hypothesize that the embeddings reside within a manifold space where Euclidean distance is not an appropriate metric for gauging the similarity between two embeddings. Meta-features are capable of grouping these embeddings into clusters, we assume each forming a sub-manifold space. When the clusters are sufficiently fine-grained, it is possible to approximate each cluster using a Gaussian distribution. Collectively, these clusters form a mixture of Gaussian distributions. By determining the posterior probabilities of the entities, the consistency of meta-features and embeddings can be assessed. 

In this study, we introduce a unique approach to evaluate the performance of pre-trained models. Specifically, we:

\begin{enumerate}
\item Adopt a distinct perspective towards the model. Instead of focusing solely on downstream performance, we emphasize the quality of the entities' representations that the model can generate.

\item Consider the features associated with the entity representations as the benchmark to assess their quality. We hypothesize that the meta-features can partition the embedding space into distinct clusters. The quality of these clusters can be evaluated using the posterior probability of Gaussian mixture models.

\item  While there are multiple methods to interpret these meta-feature spaces, we present a tree-based approach as an example that uses meta-features to segment entities into clusters.

\item Test our proposed method's effectiveness on various datasets across domains, ranging from recommendation-based to language and image models. We present both qualitative and quantitative evidence to demonstrate the approach's efficacy. 
\end{enumerate}


\section{Related Work}
This section reviews the literature on three areas: 1) Pre-trained models, 2) Vision-Language Pre-training (VLP), and 3) Pretrained Dual-Transformers (PDT) for Bipartite Graphs.

\subsection{Pre-trained models}
\textbf{Large Language Models (LLMs)}:
In recent years, significant strides have been made in the realm of Natural Language Processing (NLP), particularly with the advent of the transformer architecture. Attention-based language models such as BERT~\citep{kenton2019bert}, GPT~\citep{brown2020language}, XLNet~\citep{yang2019xlnet}, and T5~\citep{raffel2020exploring} have raised the bar in various language benchmarks. Alongside these developments, a plethora of pre-training and fine-tuning algorithms have been devised to enhance the performance of these transformer models. 
As these models grew in size, the data-driven nature and scaling characteristics of the transformer architecture became evident. These critical findings paved the way for the creation of large language models (LLMs), including LLaMa 2~\citep{touvron2023llama} with 7-70 billion parameters, BLOOM~\citep{workshop2022bloom} with 176 billion parameters, and GPT4~\citep{openai2023gpt4} with an astounding 1.7 trillion parameters.
These LLMs demonstrate impressive emergent capabilities, such as solving mathematical equations and analyzing articles, competencies not seen in prior smaller language models. These breakthroughs signify the remarkable progress made in this field.

\textbf{Vision-Language Pre-training (VLP)}: With the rapid expansion of model capacity and computational resources, the input to deep neural networks has evolved beyond a single modality, such as text or image. Vision-language pre-training (VLP) was introduced to bridge the gap between different modalities, effectively harnessing cross-modality information from both images and text.
Leveraging the successful pre-training and fine-tuning paradigm prevalent in NLP, VLP models have demonstrated exceptional performance in complex vision-language tasks. These tasks include image captioning, visual question answering, and visual reasoning.
Among the existing studies, a noteworthy contribution is the CLIP~\citep{radford2021learning} model, which employs the concept of contrastive learning to align images and text. CLIP simultaneously trains an image encoder and a text encoder on millions of image-text pairs collected from the internet. The resulting encoders have demonstrated impressive performance on downstream tasks due to their zero-shot classification capability.

\textbf{Pretrained Dual-Transformers (PDT) for Bipartite Graphs}: PDT~\citep{dai2023pdt} focuses on learning contextual knowledge from a user-content interaction dataset, which is depicted as a bipartite graph. The study identifies two key contexts in the graph: user-side and content-side. The goal of learning from these contexts is framed as two contrastive learning tasks and is applied to a recommendation task. Evaluations of two large popular datasets reveal that PDT outperforms baselines in six metrics.


\newpage

\section{Algorithm Framework}
This section presents the algorithmic framework of our proposed metric for evaluating embeddings.

For a given domain, we first collect a large size of entities with rich meta-features. Then for any given pre-trained model, we can generate an embedding dataset denoted as $\mathbf{X} = {\mathbf{x}_1,...,\mathbf{x}_N }$, where each $\mathbf{x}_i \in \mathbb{R}^d$ and $1 \leq i \leq N$. Here, $N$ represents the number of entities, and $d$ signifies the dimension of the embeddings. Simultaneously, we can produce a corresponding feature set $\mathbf{F} = {\mathbf{f}_1,...,\mathbf{f}_N }$. Each feature $\mathbf{f}_i$ comprises categorical and numerical features. We convert numerical features into categorical ones for consistency. The primary objective is to examine the consistency between these two datasets, $\mathbf{X}$ and $\mathbf{F}$.

In the simplest scenario where $\mathbf{f}_i$ has only one feature, a straightforward segmentation method is to form clusters based solely on these features, with each category creating its cluster.
However, when $\mathbf{f}_i$ has $m$ features, and each feature has $k_j$ categories, the number of combinations becomes $\prod_{j=1}^m k_j$. This results in a significantly larger number of clusters. We will postpone the discussion on finding the best split to a later session. In the upcoming discussion, we will assume that we already have a split criterion for the dataset $\mathbf{X}$.


\section{Proposed Method: Posterior Based Embedding Evaluating Metric}
 We aim to evaluate the effectiveness of different models that generate these embeddings to determine the best one. The splitting criteria, $\mathbf{S}$, can divide the entities into a group of clusters $C_1, C_2, ..., C_n$, with each entity belonging to a single cluster, where $n$ is the number of clusters. To evaluate the quality of the cluster, we adopt a posterior-based method. 

In the context of GMM, it is assumed that the data is generated from a combination of multiple Gaussian distributions. Each component of this mixture corresponds to one of the distinct clusters within the dataset. The probability of any given data point, $\mathbf{x}$,  belonging to the $k$th cluster is estimated by computing the posterior probability in the GMM framework which can be expressed as follows:

\begin{equation}
P(\theta=k|\mathbf{x}) = \frac{P(\mathbf{x}|\theta = k) P(\theta = k)}{\sum_{j=1}^m P(\mathbf{x}|\theta=j)P(\theta=j)},
\end{equation}

where $P(\theta = k)$ = $\frac{\text{number of points in the $k$th cluster  }}{N}$


To assess the quality of the embeddings $\mathbf{X}$ within the context of a splitting $\mathbf{S}$, we compute the overall evaluation metric by averaging the log probabilities of all the embeddings across all clusters. This metric provides an assessment of the quality of the embeddings. We call it the average of the log posterior ($ALP$).

\begin{equation}
ALP_\mathbf{S}^\mathbf{X} = \frac{1}{N}\sum_{k=1}^m \sum_{\mathbf{x}_i \in C_k} \log P(\theta=k|\mathbf{x}_i) 
\label{eq:PosterEqn}
\end{equation}

One challenge with the formula above is its high sensitivity to outliers. A single outlier could lead to an extremely large value for $ALP_\mathbf{S}^\mathbf{X}$. We implement a clipping mechanism for embeddings with very small posterior probabilities to mitigate the impact of such outlier entities. Specifically, if $P(\theta=k|\mathbf{x}_k)$ is less than $k/N * \varepsilon$, we exclude the entity from the $ALP_\mathbf{S}^\mathbf{X}$ computation.

Another challenge arises when the embeddings exist in large dimensions. If the number of embeddings in each cluster is smaller than the embedding dimension, this leads to a rank-deficient covariance. To address this, we propose a multi-head solution. In this approach, each head is a randomly selected $v$ dimensions from the existing $d$ dimensions and $ALP$ is estimated based on these dimensions. This process is repeated multiple times, and the average results are used which we refer to as the Mean of the . This concept is inspired by the random forest algorithm~\citep{breiman2001random} and Matryoshka Representation Learning \citep{Kusupati2022MatryoshkaRL}.     Additionally, we apply the routine regularization approach, i.e., adding $\epsilon$$I$ to the covariance matrix. The value of $\epsilon$ is decided in the following manner.

\begin{equation}
    \epsilon = max(\lambda_k/(10 D) \lambda_0, 1e{-}8),
\end{equation}

Where $D$ is the dimensionality of the embeddings and $\lambda_i$ are the eigenvalues of the covariance matrix (sorted decreasingly by their magnitude). $k$ is the minimum value that satisfies $\frac{\Sigma_{i=0}^k\lambda_i}{\Sigma_{i=0}^D\lambda_i}>99.99\%$.

\subsection{One Meta Feature Based Clustering}
In the simplest scenario, where the feature vector $\mathbf{f}_i$ consists of only one feature, a straightforward and intuitive approach to segmentation is to form clusters based solely on these features. This method is straightforward as it capitalizes on the inherent characteristics of the data. Each unique category within the data forms its distinct cluster, effectively grouping similar items. The consistency of these clusters with the embeddings can serve as a measure of the quality of the embeddings. However, it's important to note that extending this approach to accommodate more than two meta-features is not as straightforward.

\subsection{Meta features + representation based segmentation}
Inspired by \textit{EmbeddingTree} algorithm, we can construct the tree based on the entities, and all the leaf nodes are the final clusters, specifically:
We first convert non-binary categorical features into binary ones by asking yes-no questions regarding each of their categorical values and get the binary feature sets: $\mathbf{G} = \{\mathbf{g}_1,...,\mathbf{g}_N \}$ ($\mathbf{g}_i \in \{0,1\}^q$, $1 \leq  i \leq N$), $q$ denote the total number of converted binary features.  

With the processed data, we describe the \textit{EmbeddingTree} algorithm with details in Algorithm \ref{ET}.
We iterate through the $q$ features (line 6) and evaluate them based on the splitting criteria described in Section \ref{split} to pick out the best feature for splitting (line 8-10), using the feature's binary value (line 11-13).
The above procedure is executed recursively (lines 15-16) until the splitting criterion ($\Theta$), e.g., the number of entities per tree node or the tree depth, is no longer satisfied (line 2).
With the given embedding and feature data, the whole procedure is deterministic. 


\begin{algorithm}[h]
  \caption{Build an \textit{EmbeddingTree}}\label{ET}
  \footnotesize
   \begin{algorithmic}[1]
   \Procedure{BuildTree}{$[\mathbf{X},\mathbf{F}], q, \Theta$}
    \If{$\Theta \hspace{1mm} is \hspace{1mm} not\hspace{1mm} satisfied$ }
    \State return LeafNode([$\mathbf{X},\mathbf{F}$])
    \Else
    \State $max\_t \leftarrow -\infty$ 
        \For{$k \in \{1,..., q\}$}
        \State $t = Embedding- MAP([\mathbf{X}, \mathbf{F}^{k}])$
            \If{$t > max\_t$}
            \State $bestFea = k$ 
            \State $max\_t = t$
            \EndIf
        \EndFor
    \EndIf
    \State $[\mathbf{X},\mathbf{F}]_{left} = \{\mathbf{x} \in \mathbf{X} \vert  \mathbf{F}_{bestFea} == 0\} $\\
    \State $[\mathbf{X},\mathbf{F}]_{right} = \{\mathbf{x} \in \mathbf{X} \vert  \mathbf{F}_{bestFea} == 1\} $\\
    \State Children.Left = $BuildTree([\mathbf{X},\mathbf{F}]_{left}, q, \Theta) $
    \State Children.Right = $BuildTree([\mathbf{X},\mathbf{F}]_{right}, q, \Theta) $
    \State \Return Children
   \EndProcedure
 \end{algorithmic}
\end{algorithm}

\subsection{2-GMM Splitting with Maximum A Posteriori Estimation (MAP)}
\label{split}
One critical component of Algorithm~\ref{ET} is the criterion for selecting the best splitting feature. The criterion is computed based on the approximate MAP for GMMs inspired by~\citep{zheng2023embeddingtree}.

We assume the embedding can be modeled as two mixture Gaussians. The expectation-maximization (EM) algorithm is used to estimate all the parameters and latent variables jointly. The latent variables, $z_{i,j}$, denote the probability that sample $i$ is in cluster $j$.
With $N$ as the number of observations and $J$ as the number of Gaussian clusters (in this case, $J = 2$), $z = \{z_{1,1}, z_{1,2}, . . . ,z_{N, J-1}, z_{N, J} \}$, the complete likelihood (including the latent variables) is:
\begin{equation}
P(\mathbf{x},\mu,\Sigma,w,z )= \prod_{i=1}^N \prod_{j=1}^J \{w_j \mathcal{N}(\mathbf{x}_n;\mu_j,\Sigma_j^2)\}^{z_{i,j}},
\end{equation}
where $\mu$ is the mean vectors and $\Sigma$ is covariance matrix of the Gaussians.

We go through every feature to find the best binary feature that splits the embedding and forms the best GMM. Each candidate binary feature splits the embeddings into two clusters, each formulated as a Gaussian. For each feature, suppose the first $s$ embeddings have feature value $ Fk=0$ and the rest $N-s$ embeddings have feature value $ Fk=1$. 
We estimate both clusters' weights, means, and variances using the maximum likelihood estimation (MLE).

\[
\hat{\mu_1} = \frac{1}{s}\sum_{i = 1}^s \mathbf{\mathbf{x}_i},\hspace{5mm}
\hat{\Sigma_1} = \frac{1}{s} \sum_{i=1}^s \mathbf{(\mathbf{x}_i - \hat \mu_1) (\mathbf{x}_i -\hat  \mu_1)}^T,
\hspace{5mm}
\hat{w_1}= \frac{s}{N},
\]
\[
\hat{\mu_2} = \frac{1}{N-s}\sum_{i=s+1}^N \mathbf{\mathbf{x}_i},\hspace{5mm}
\hat{\Sigma_2} = \frac{1}{N-s} \sum_{i=s+1}^{N} \mathbf{(\mathbf{x}_i - \hat \mu_2) (\mathbf{x}_i -\hat  \mu_2)}^T,
\]
\[
\hat{w_2}= \frac{N-s}{N}.
\]

In other words, our algorithm performs a hard clustering rather than the soft clustering of GMM. Thus, if $\mathbf{x}_i$ is in cluster $j$, then $z_{i,j} = 1$ and $z_{i,j'} = 0$ for all $j \neq j'$. Given this approximation, the likelihood can be obtained by summing over the $z$:
\begin{equation}
P(\mathbf{x},\mu,\Sigma,w)= \sum_z \prod_{i=1}^N \prod_{j=1}^J \{w_j \mathcal{N}(\mathbf{x}_n;\mu_j,\Sigma_j^2)\}^{z_{i,j}}
\end{equation}

Note that $z_{(i\in(0,s],j=1)}=z_{(i\in[s+1,N),j=2)}=1$ and $z_{i,j}=0$, otherwise, the above equation can be simplified to:

\begin{equation}
P(\mathbf{x},\mu,\Sigma,w )= \prod_{i=1}^s w_1 \mathcal{N}(\mathbf{x}_i;\mu_1,{\Sigma_1}^2)\prod_{i=s+1}^N w_2 \mathcal{N}(\mathbf{x}_i;\mu_2,{\Sigma_2}^2).
\end{equation}

We can treat each split feature as another random variable $\theta$. To choose the best-split feature, we want to maximize $P(\mathbf{x},\mu,\Sigma,w, \theta)$; in other words, we want to find $\theta$ that gives the largest $P(\mathbf{x},\mu,\Sigma,w )$.

\subsection{Finding the Best Splitting Point}
For simplicity, we only consider $\theta$ as the random variable we want to estimate; by injecting the prior information into the formula, we can treat each splitting feature with a different weight. By applying Maximum A Posteriori Estimation (MAP), we can formulate the problem as follows:
\begin{equation}
P(\theta_i|\mathbf{x}) = \frac{P(\mathbf{x}|\theta_i) P(\theta_i)}{\sum_{j=1}^q P(\mathbf{x}|\theta_j)P(\theta_j)},
\end{equation}
where $q$ is the number of possible splits.

By plugging (3) into (4), we can get
\begin{equation}
P(\theta_i|\mathbf{x}) = \frac{\prod_{k=1}^s w_1 \mathcal{N}(\mathbf{x}_k;\mu_1,{\Sigma_1}^2, \theta_i)\prod_{k=s+1}^N w_2 \mathcal{N}(\mathbf{x}_k;\mu_2,{\Sigma_2}^2, \theta_i) p(\theta_i)}{\sum_{j=1}^q \prod_{k=1}^s w_1 \mathcal{N}(\mathbf{x}_k;\mu_1,{\Sigma_1}^2,\theta_j)\prod_{k=s+1}^N w_2 \mathcal{N}(\mathbf{x}_k;\mu_2,{\Sigma_2}^2, \theta_j)p(\theta_j)}.
\end{equation}

Plugging in the estimates for all the parameters and taking the log of $P(\theta_i|\mathbf{x})$, we can get
\begin{multline}
    \log \hat{P}(\theta_i | \mathbf{x}) = \sum_{i=1}^s[\log\hat{w_1}+\log\mathcal{N}(\mathbf{x}_i;\hat{\mu_1},\hat{\Sigma_1}^2)]
   +\sum_{i=s+1}^N[\log\hat{w}_2+\log\mathcal{N}(\mathbf{x}_i;\hat{\mu_2},\hat{\Sigma_2}^2)] + \log p(\theta_i) \\ - \log(\sum_{j=1}^q \prod_{k=1}^s w_1 \mathcal{N}(\mathbf{x}_k;\hat{\mu}_1,{\hat{\Sigma}_1}^2,\theta_j)\prod_{k=s+1}^N w_2 \mathcal{N}(\mathbf{x}_k;\hat{\mu}_2,{\hat{\Sigma}_2}^2, \theta_j)p(\theta_j)).
\end{multline}

By applying this formula, we can use the prior knowledge of the importance of the feature to find the split that maximizes $\log \hat{P} $.







\subsection{Embedding Comparison based on the same splitting criteria}
If we have two sets of embeddings, $\mathbf{X_A} = \{\mathbf{x_A}_1, \ldots, \mathbf{x_A}_N \}$, and $\mathbf{X_B} = \{\mathbf{x_B}_1, \ldots, \mathbf{x_B}_N \}$, both trained on the same dataset but using different models, denoted as models $A$ and $B$, where ($\mathbf{x_A}_i, \mathbf{x_B}_i \in \mathbb{R}^p$, $1 \leq i \leq N$), we can generate two corresponding splitting criteria, $\mathbf{S_A}$ and $\mathbf{S_B}$. 
The objective now is to assess and compare the quality of these two sets of embeddings. Let's represent $ALP_{\mathbf{S}^\mathbf{A}}^{\mathbf{X}^\mathbf{A}}$ as $ALP_\mathbf{A}^\mathbf{A}$ for embeddings $\mathbf{X}_\mathbf{A}$ and splitting criteria $\mathbf{S}_\mathbf{A}$. Given two sets of embeddings, $\mathbf{X_\mathbf{A}}$ and $\mathbf{X_\mathbf{B}}$, along with two corresponding splitting criteria, $\mathbf{S_\mathbf{A}}$ and $\mathbf{S_B}$, we can define four metrics: $ALP_\mathbf{A}^\mathbf{A}$, $ALP_\mathbf{B}^\mathbf{B}$, $ALP_\mathbf{A}^\mathbf{B}$, and $ALP_\mathbf{B}^\mathbf{A}$. We need to fix the splitting criteria to do clustering, so a proper comparison should be between $ALP_\mathbf{A}^\mathbf{A}$ and $ALP_\mathbf{A}^\mathbf{B}$ or between $ALP_\mathbf{B}^\mathbf{A}$ and $ALP_\mathbf{B}^\mathbf{B}$.


\section{Experimental Analysis}
\label{sec:exp}

In this experimental session, we initially conducted experiments on a synthetic dataset to verify the effectiveness of our proposed algorithm. Following this, we further evaluate the results in three areas: the MovieLens dataset~\citep{harper2015movielens} for relational models, spatial datasets~\citep{gurnee2023language} for large language models, and the Robustness library \citep{robustness,engstrom2019adversarial,santurkar2019image,santurkar2020breeds} for image models.

\subsection{Synthetic Dataset: Gaussian Mixture Model (GMM) of Ten Gaussian Distributions}
\label{sec:Syntdata}

\begin{figure}[H] 
    \centering
    \begin{subfigure}[b]{0.32\textwidth}
        \includegraphics[width=\textwidth]{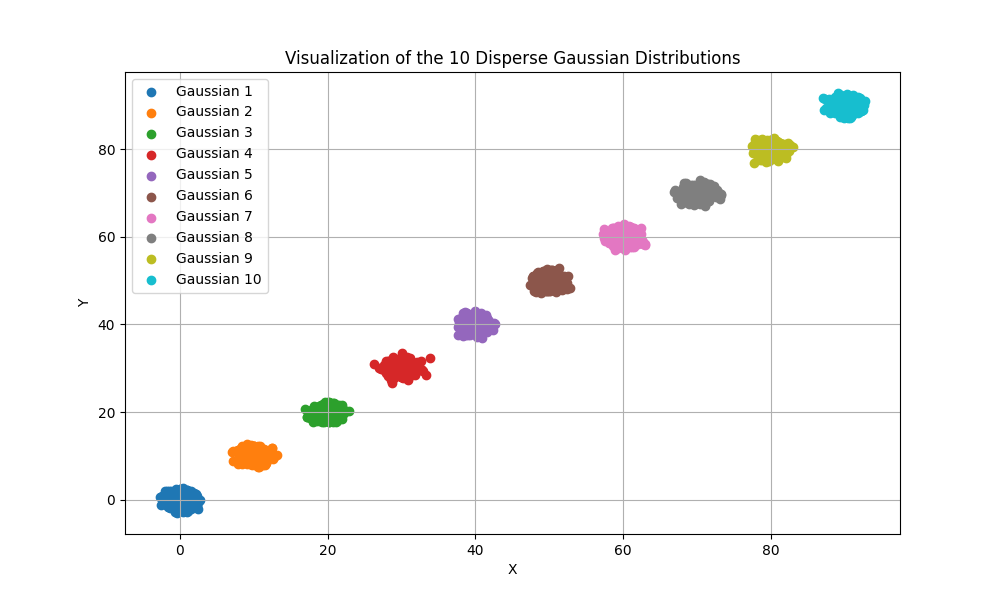}
        \caption{Perfectly separated clusters}
        \label{fig:first_plot}
    \end{subfigure}
    \hfill 
    \begin{subfigure}[b]{0.32\textwidth}
        \includegraphics[width=\textwidth]{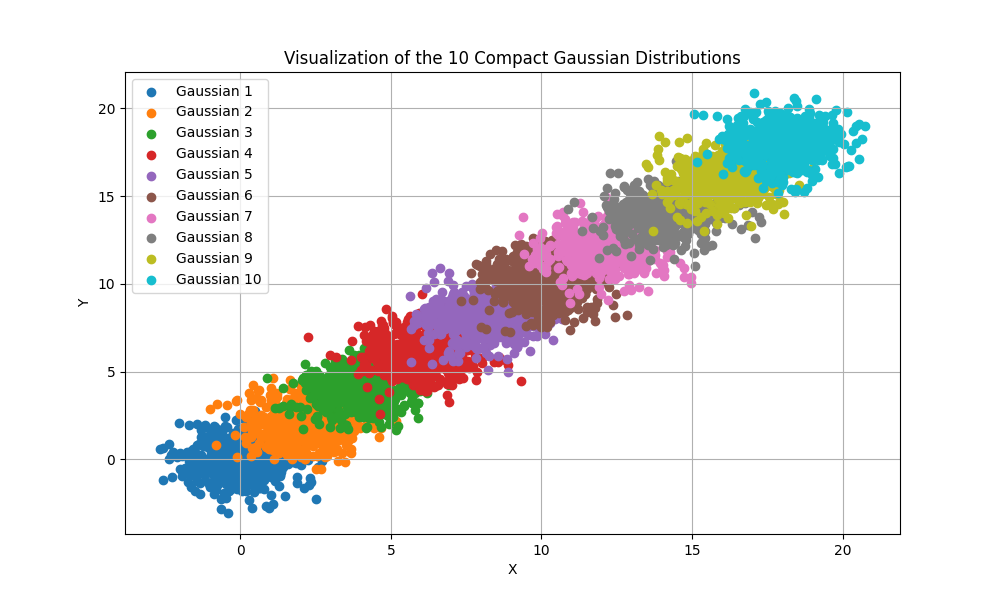}
        \caption{Partially overlapping clusters}
        \label{fig:second_plot}
    \end{subfigure}
    \hfill
    \begin{subfigure}[b]{0.32\textwidth}
        \includegraphics[width=\textwidth]{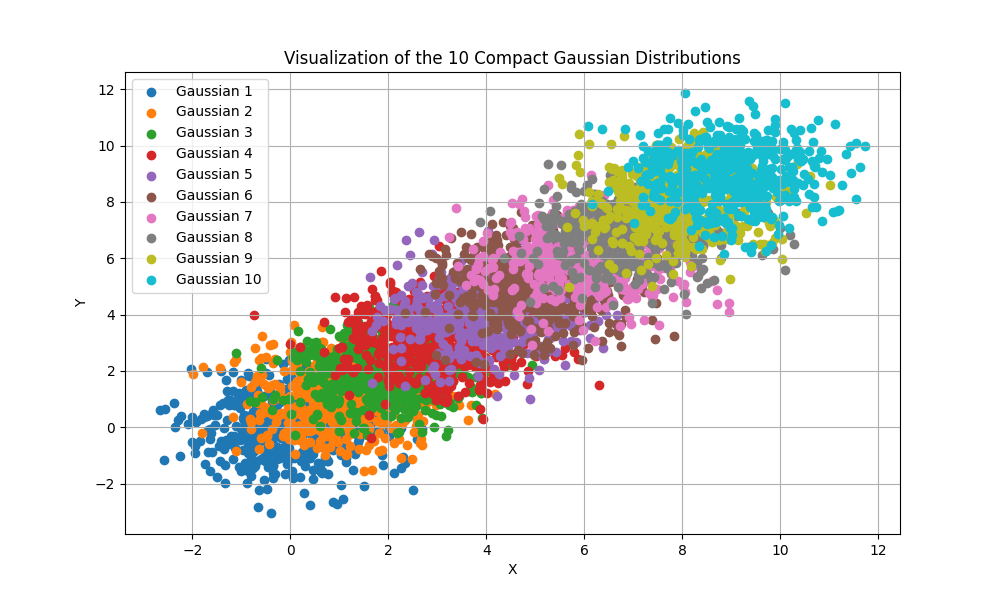}
        \caption{Perfectly overlapping clusters}
        \label{fig:third_plot}
    \end{subfigure}
    \caption{Illustration on a 2D synthetic dataset consisting of 10 Gaussian distributions that are perfectly separated, partially overlapping, and perfectly overlapping.}
    \label{fig:GMM}
\end{figure}

To validate the efficacy of our proposed posterior-based embedding evaluation metric, as outlined in \autoref{eq:PosterEqn}, we designed an experiment encompassing three scenarios, each with $10$ clusters. These clusters were generated such that they are either perfectly separated, partially overlapping, or perfectly overlapping and are all generated using a Gaussian Mixture Model. \autoref{fig:GMM} presents the results from these scenarios. As anticipated, the Average of the Log Posterior (ALP) scores for the ten (10) perfectly separated Gaussian Distributions was $0$, and the accuracy of the clusters assigned from the posterior matrix was $100\%$. In the case of 10 partially overlapping Gaussian Distributions, the ALP score was $-0.3285$, and the accuracy of the clusters assigned from the posterior matrix was $86.96\%$. Similarly, for the ten (10) perfectly overlapping Gaussian Distributions, the ALP score was $-0.9372$, and the accuracy of cluster assignment was $57.34\%$.


\subsection{Moive Lens Dataset for Relational }
MovieLens-25M consists of 25,000,095 reviews made by 162,541 reviewers on 59,047 movies. We compare the following models to generate the movie embeddings. Word2vec \citep{Mikolov2013EfficientEO}, PDT \citep{dai2023pdt} and SASRec~\citep{SASRec}. From the Word2Vec model, we generate two distinct types of embedding representations, specifically \texttt{w2v\_single} and \texttt{w2v\_combine}. In the case of \texttt{w2v\_single}, we create a sequence of movies for each reviewer's review, sort it based on time, and then employ the Word2vec model to learn the movie embeddings.
On the other hand, \texttt{w2v\_combine} not only includes the sequences of movies for reviewers but also incorporates sequences of reviewers for each movie and reviewer/movie pairs as part of the training data. 
For both \texttt{SASRec} and \texttt{PDT}, we generate sequences of nine movies for each reviewer. Additionally, for \texttt{PDT}, we generate sequences of reviewers for each movie, as this is a necessary input for \texttt{PDT}. \texttt{SASRec} is trained using BPR loss, while \texttt{PDT} is trained using two contrastive losses.
Both \texttt{PDT}  and \texttt{SASRec} are contextualized embeddings, while \texttt{w2v\_single} and  \texttt{w2v\_combine} are static embeddings.
We employ two clustering techniques. The first approach involves clustering by single meta-features, such as year and genre. We also apply the Embedding tree-based method to generate the tree for both year and genre features and use the leaf nodes as clusters.

\subsubsection{Movie Lens Dataset: Clustering by Year}

\begin{figure*}[htp]
    \centering
    \begin{subfigure}[b]{0.3\textwidth}
        \includegraphics[width=\textwidth]{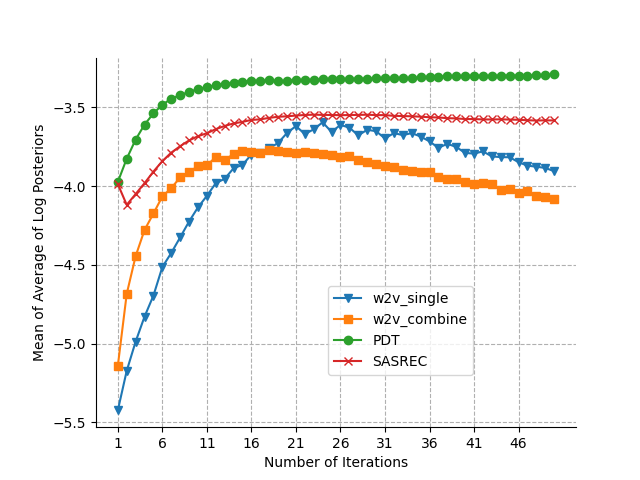}
        \caption{Mean of the average of log posteriors}
    \end{subfigure}
    \hspace{2cm} 
    \begin{subfigure}[b]{0.3\textwidth}
        \includegraphics[width=\textwidth]{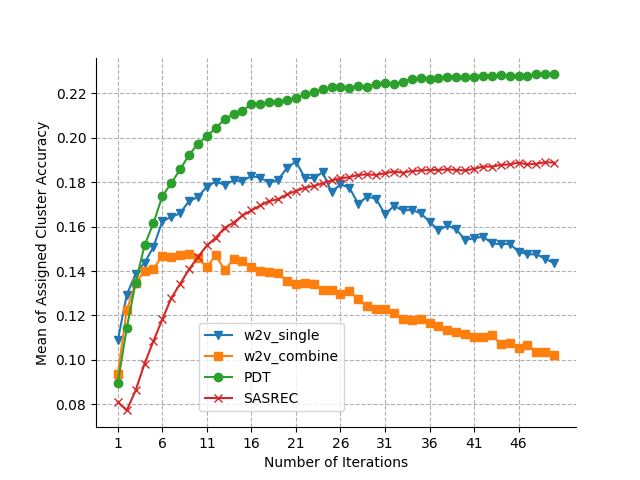}
        \caption{Mean of accuracy of assigned clusters}
    \end{subfigure}
    \caption{Mean of the average of the log posterior and accuracy on the MovieLens dataset by clustering on year.}
    \label{fig:YearClusters}
\end{figure*}

We evaluated and compared four kinds of embedding representations trained on the MovieLens Dataset. These were trained at various iteration levels; we use the ``year'' feature as labels to cluster the embeddings. As illustrated \autoref{fig:YearClusters} (a), the \texttt{PDT} and \texttt{SRSREC} embedding performed better than all embeddings across all iterations, as seen in the Mean of the average of log posteriors plot. During the early stages of training from iteration 1 to 16, \texttt{w2v\_combine}  outperformed \texttt{w2v\_single}. However, in the later stages from iteration 16 to 50, \texttt{w2v\_single} superseded \texttt{w2v\_combine}.

As depicted in the Mean of accuracy of assigned clusters plot of \autoref{fig:YearClusters} (b), \texttt{PDT} and \texttt{SRSREC} demonstrates a consistent and stable performance over all other types of embeddings across all iterations. Generally, \texttt{w2v\_single} exceeds the performance of \texttt{w2v\_combine}. This suggests that contextualized embeddings, specifically \texttt{PDT} and \texttt{SRSREC}, most effectively encode year information and remain stable across all iterations. Also, \texttt{w2v\_single} demonstrates superior encoding of year information compared to \texttt{w2v\_combine}.

\subsubsection{Movie Lens Dataset: Clustering by Genre}
\label{sec:MovieLensGenre}

\begin{figure*}[htp]
    \centering
    \begin{subfigure}[b]{0.3\textwidth}
        \includegraphics[width=\textwidth]{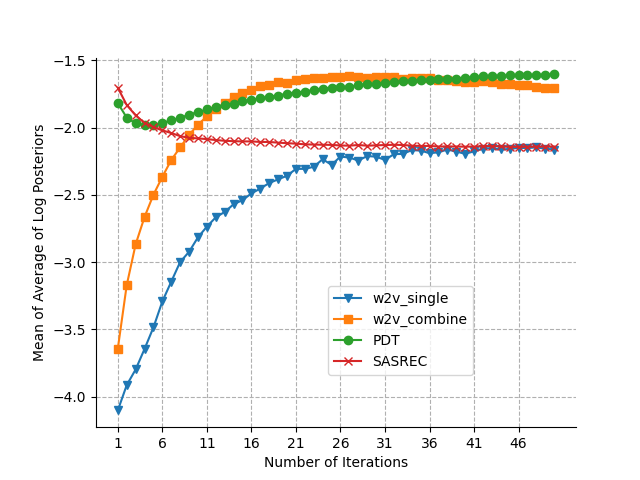}
        \caption{Mean of the average of log posteriors}
    \end{subfigure}
    \hspace{2cm} 
    \begin{subfigure}[b]{0.3\textwidth}
        \includegraphics[width=\textwidth]{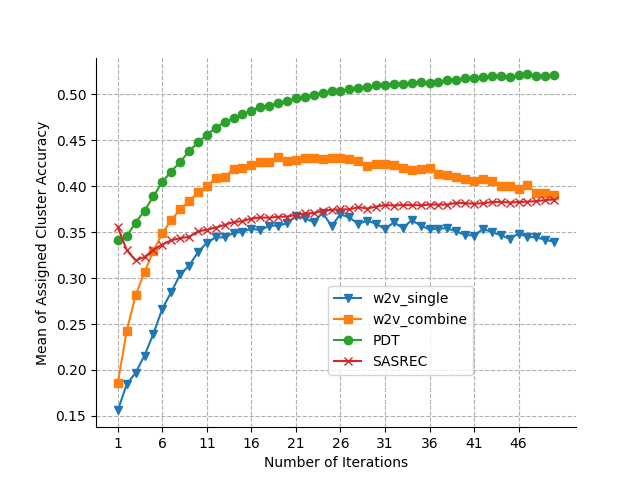}
        \caption{Mean of accuracy of assigned clusters}
    \end{subfigure}
    \caption{Mean of the average of the log posterior and accuracy on the MovieLens dataset by clustering on the genre.}
    \label{fig:GenreClusters}
\end{figure*}

Here, we created clusters with the genre features as labels. We then compute and report the Mean of the average of log posteriors and the Mean of accuracy of assigned clusters. These findings are presented in \autoref{fig:GenreClusters}. Contrary to the consistent pattern observed with year features as labels, the genre features do not exhibit a similar consistency. From \autoref{fig:GenreClusters} (a), it's noticeable that the \texttt{PDT} embedding generally outperforms both \texttt{SASRec} and \texttt{w2v\_single} over all iterations. Furthermore, \texttt{SASRec} surpasses \texttt{w2v\_single} from the 1st to the 40th iteration, after which they both plateau with similar scores. Between the 12th and 36th iterations, \texttt{w2v\_combine} is observed to outperform \texttt{PDT}. Moving to the Mean accuracy of the assigned clusters plot (\autoref{fig:GenreClusters} (b)), it's evident that \texttt{PDT} consistently outperforms all other embedding types across all iterations. Generally, \texttt{w2v\_combine} surpasses both \texttt{SASRec} and \texttt{w2v\_single}, except for the first and third iterations where \texttt{SASRec} exceeds \texttt{w2v\_combine}.

\subsubsection{Movie Lens Dataset: Clustering with Tree Leaf Nodes on Genre and Year as the Meta Features}
\label{sec:MovieLensYearMeta}

\begin{figure*}[htp]
    \centering
    \begin{subfigure}[b]{0.3\textwidth}
        \includegraphics[width=\textwidth]{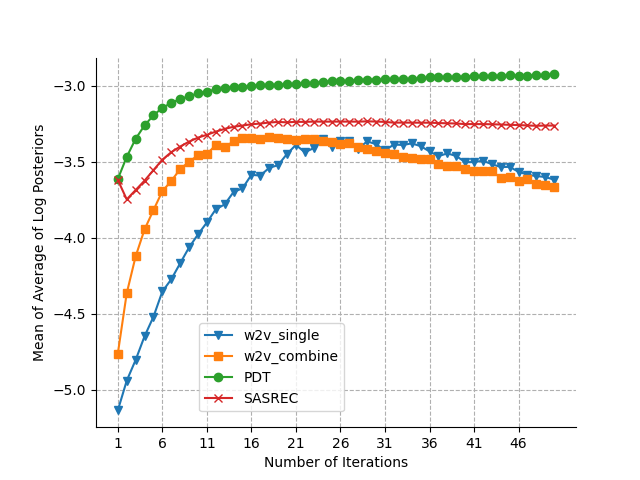}
        \caption{Mean of the average of log posteriors}
    \end{subfigure}
    \hspace{2cm} 
    \begin{subfigure}[b]{0.3\textwidth}
        \includegraphics[width=\textwidth]{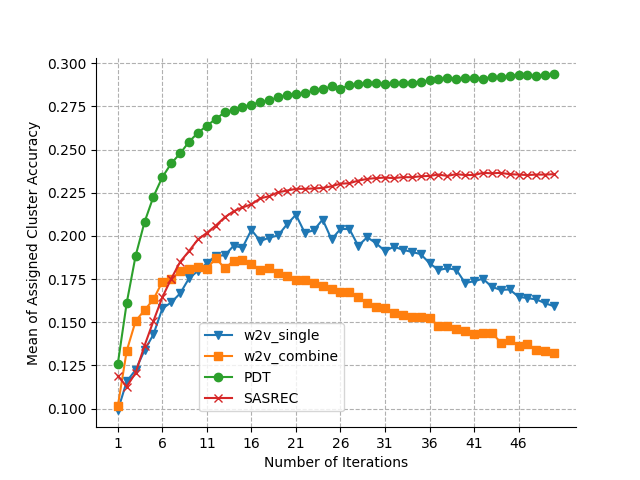}
        \caption{Mean of accuracy of assigned clusters}
    \end{subfigure}
    \caption{Mean of the average of the log posterior and accuracy on the MovieLens dataset by clustering with tree leaf nodes.}
    \label{fig:GenreYearTreeClusters}
\end{figure*}

We also explored the case where we used the tree leaf nodes from the embedding tree constructed with year and genre as meta-features. The Mean average of log posteriors of assigned clusters is calculated and reported, as shown in \autoref{fig:GenreYearTreeClusters} (a). The \texttt{PDT} and \texttt{SASRec} embeddings consistently surpass other embeddings throughout all iterations. However, we notice that \texttt{w2v\_combine} surpasses \texttt{w2v\_single} from the 1st to the 26th iteration, but \texttt{w2v\_single} overtakes \texttt{w2v\_combine} from the 26th to the 50th iteration. The Mean accuracy of assigned clusters, illustrated in \autoref{fig:GenreYearTreeClusters} (b), clearly shows that the \texttt{PDT} and \texttt{SASRec} embedding exhibit a steady and consistent increase in performance compared to all other embeddings across all iterations. This is followed by \texttt{w2v\_single}, which generally surpasses \texttt{w2v\_combine}.

All the above experiment results suggest that the contextualized embeddings \texttt{SASRec} and \texttt{PDT} are more effective at capturing the semantic and structural relationships in the input data compared to their static embedding counterparts, namely \texttt{w2v\_single} and \texttt{w2v\_combine}, which meet our expectations.

\subsection{Experiments with Embedding from Llama-2 Models}

In this session, we implement the proposed concept in the field of large language models and use Llama-2 as an example.

\textbf{Models}: The released Llama-2 models (\cite{touvron2023llama}) have three versions with different model sizes and embedding dimensionalities. These details have been included in \autoref{tab:llama2}.

\textbf{Datasets}: We use the Llama-2 models to generate embeddings for three spatial datasets introduced by~\citep{gurnee2023language}. The details of these datasets are shown in \autoref{tab:mapdata}. Each dataset instance is a word/phrase for a location (e.g., the name of a city, a university, or a place of interest). Based on the spatial location as a meta feature, we generate the clusters for each dataset, i.e., based on the continent, state, and borough of the instances from the \textit{World Place}, \textit{USA Place}, and \textit{NYC Place} dataset, respectively.

\begin{table*}[hb]
\begin{minipage}{0.45\textwidth}
\centering
\caption{Details of the Llama-2 models.}
\label{tab:llama2}
\begin{tabular}{lccc}
\toprule
Model       & \#head & \#layer & \#dim \\
\midrule
Llama-2-7b  & 32     & 32      & 4096 \\
Llama-2-13b & 40     & 40      & 5120 \\
Llama-2-70b & 64     & 80      & 8192 \\
\bottomrule
\end{tabular}
\end{minipage}%
\hspace{0.01\textwidth} 
\begin{minipage}{0.45\textwidth}
\centering
\caption{Details of the three spatial datasets.}
\label{tab:mapdata}
\begin{tabular}{lccc}
\toprule
Dataset     & \#clusters & \#samples & Cluster   \\
\midrule
World Place & 8          & 39585     & continent \\
US Place    & 49         & 29997     & state     \\
NYC Place   & 7          & 19838     & borough   \\
\bottomrule
\end{tabular}
\end{minipage}
\hspace*{\fill} 
\end{table*}

\textbf{Embedding Quality}: We feed the three datasets into the three (3) Llama-2 models and get their activations/embeddings at each layer of the models. Specifically, we obtain $32$, $40$, and $80$ sets of embeddings for the three sets of models (as the three models have $32$, $40$, and $80$ layers, respectively). We use our proposed method for each set of embeddings to compute the posterior of individual instances falling into the correct cluster.

\begin{figure*}[ht]
\centering
\includegraphics[width=0.8\textwidth]{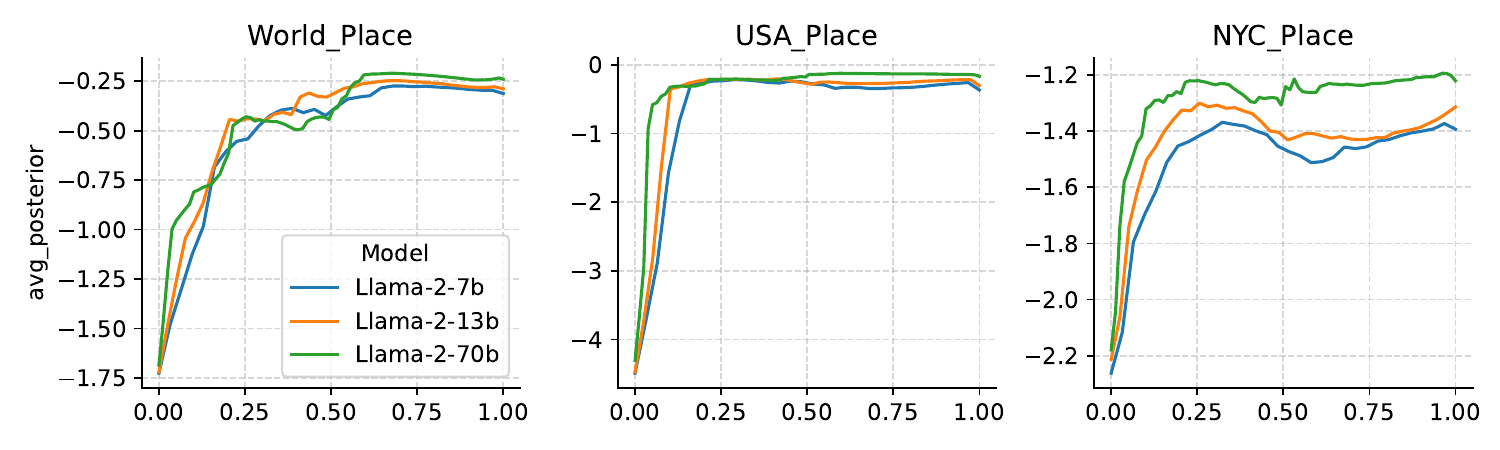} 
\caption{Embedding quality over model layers (average of the log posterior). The dimensions are divided into subsets, each comprising 128 dimensions.}
\label{fig:llama2_multihead_avg}
\end{figure*}

\begin{figure*}[ht]
\centering
\includegraphics[width=0.8\textwidth]{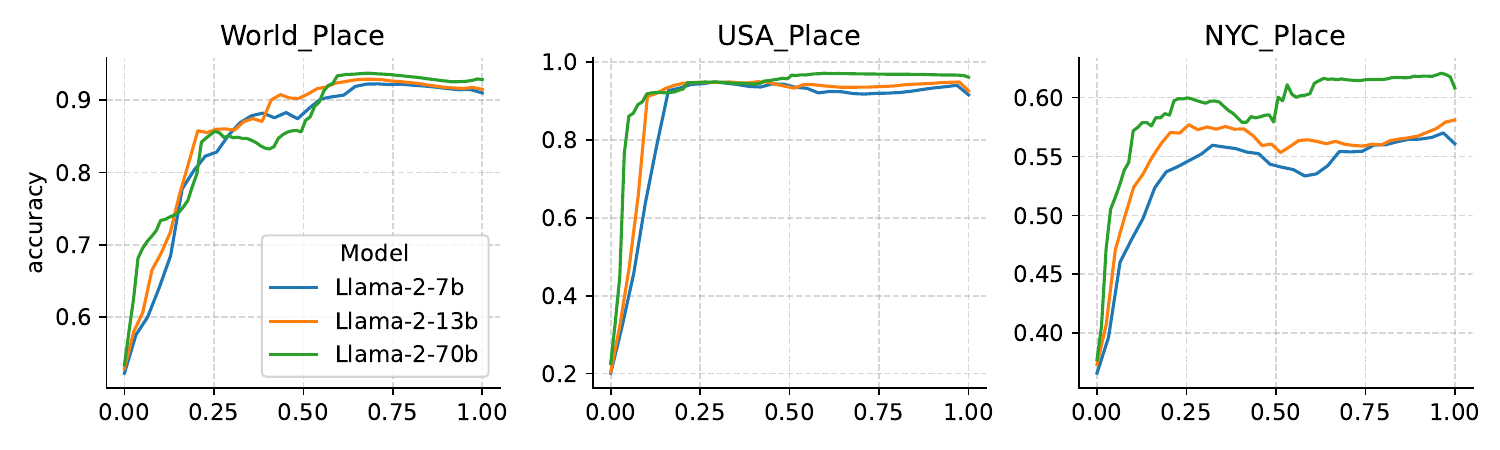} 
\caption{Embedding quality over model layers (accuracy of assigned clusters). The dimensions are divided into subsets, each comprising 128 dimensions.}
\label{fig:llama2_multihead_acc}
\end{figure*}

From \autoref{fig:llama2_multihead_avg} and \autoref{fig:llama2_multihead_acc},  we present the effectiveness of embeddings across three different datasets and models. The x-axis of each graph indicates the percentage of layers relevant to the corresponding Llama-2 models, while the y-axis represents the Mean average of log posteriors and accuracy. The results show a noticeable upward trend in the quality of embeddings from the initial to the final layers. Specifically, in the World\_Place, USA\_Place, and NYC\_Place datasets, the green lines, which denote the larger Llama-2-70b model, exhibit the highest levels of posteriors and accuracy. This indicates that the Llama-2-70b model is more adept at incorporating location data compared to the other models.

\subsection{Experiment analysis of CLIP Embeddings on the BREEDS hierarchy}
\label{ssec:CLIPBreedsPerData}
In this section, we assess the ability of our metric to predict the classification performance on embeddings of 3 pre-trained CLIP \citep{radford2021learning} models on datasets from the Breeds Hierarchy \citep{santurkar2020breeds}, namely datasets entity 13, entity 30, living 17 and nonliving 26. Next, we briefly describe the Breeds Dataset and the CLIP models used for our evaluation.

\textbf{Dataset: }Breeds Hierarchy \citep{santurkar2020breeds} was created by leveraging and modifying the WordNet \citep{miller-1994-wordnet} hierarchy for the ImageNet \citep{5206848} dataset to group together semantically similar classes into one (1) superclass. The original purpose of creating this hierarchy was to use the subpopulations present in superclasses to detect a model's robustness to distribution shift. For this experiment, we leverage the entire dataset to evaluate the performance of our metric in predicting the generalization of CLIP models on those embeddings.

\textbf{Models: }For this study, we use CLIP-ViT transformers ViT-L/14, ViT-B/32, and ViT-B/16 trained on 224 px $\times$ 224 px images as image encoders, where ViT-L/14 encodes each image into a 768-dimensional embedding and ViT-B/16 and ViT-B/32 encode each image as a 512-dimensional embedding. We obtained model embeddings for each network from the final image encoder and the mean embeddings of internal multi-head self-attention layers. We train a linear probe on the ImageNet \citep{5206848} training subset of the Breeds dataset and learn the parameters for estimating the log posterior (ALP). To compute the average log posterior, we don't split the embeddings into multiple blocks; therefore, the average log posterior (ALP) demonstrates the flexibility of our approach. We correlate the performance of these two metrics to understand our trained posterior's behavior better and present the analysis and results next.

\begin{table}[htb]
\centering
\caption{Pearson's Correlation on Breeds Datasets (Part 1): Comparing layerwise log posterior and linear probe accuracy across regularization levels in the training and validation sets of the Breeds Hierarchy.}
\label{tab:breedsCLIPpearcorr1}
\begin{minipage}{.45\linewidth}
    \centering
    \caption*{Entity 13 Dataset}
    \setlength{\tabcolsep}{3pt}
    \scalebox{0.8}{
    \begin{tabular}{@{}lcccccc@{}}
        \toprule
        & \multicolumn{3}{c}{Train Set} & \multicolumn{3}{c}{Val. Set} \\
        \cmidrule(lr){2-4} \cmidrule(lr){5-7}
        Reg. & B16 & B32 & L14 & B16 & B32 & L14 \\
        \midrule
        $Diag$ & .99 & .99 & .98 & .99 & .99 & .97 \\
        $10^{-3}$ & .97 & .99 & .97 & .97 & .98 & .95 \\
        $10^{-6}$ & .97 & .96 & .98 & .99 & .98 & .99 \\
        $10^{-9}$ & .96 & .95 & .98 & .99 & .99 & .99 \\
        \bottomrule
    \end{tabular}
    }
\end{minipage}
\begin{minipage}{.45\linewidth}
    \centering
    \caption*{Entity 30 Dataset}
    \scalebox{0.8}{
    \begin{tabular}{@{}lcccccc@{}}
        \toprule
        & \multicolumn{3}{c}{Train Set} & \multicolumn{3}{c}{Val. Set} \\
        \cmidrule(lr){2-4} \cmidrule(lr){5-7}
        Reg. & B16 & B32 & L14 & B16 & B32 & L14 \\
        \midrule
        $Diag$ & .99 & .99 & .99 & .99 & .99 & .98 \\
        $10^{-3}$ & .98 & .97 & .98 & .98 & .98 & .98 \\
        $10^{-6}$ & .85 & .83 & .89 & .97 & .96 & .99 \\
        $10^{-9}$ & .8 & .76 & .87 & .96 & .96 & .98 \\
        \bottomrule
    \end{tabular}
    }
\end{minipage}
\end{table}

\begin{table}[htb]
\centering
\caption{Pearson's Correlation on Breeds Datasets (Part 2): Comparing layerwise log posterior and linear probe accuracy across regularization levels in the training and validation sets of the Breeds Hierarchy.}
\label{tab:breedsCLIPpearcorr2}
\begin{minipage}{.45\linewidth}
    \centering
    \caption*{Living 17 Dataset}
    \setlength{\tabcolsep}{3pt}
    \scalebox{0.8}{
    \begin{tabular}{@{}lcccccc@{}}
        \toprule
        & \multicolumn{3}{c}{Train Set} & \multicolumn{3}{c}{Val. Set} \\
        \cmidrule(lr){2-4} \cmidrule(lr){5-7}
        Reg. & B16 & B32 & L14 & B16 & B32 & L14 \\
        \midrule
        $Diag$ & .99 & .99 & .99 & .99 & .99 & .98 \\
        $10^{-3}$ & .93 & .93 & .95 & .97 & .96 & .99 \\
        $10^{-6}$ & .68 & .65 & .72 & .93 & .94 & .97 \\
        $10^{-9}$ & .57 & .49 & .64 & .93 & .95 & .97 \\
        \bottomrule
    \end{tabular}
    }
\end{minipage}%
\begin{minipage}{.45\linewidth}
    \centering
    \caption*{Non Living 26 Dataset}
    \scalebox{0.8}{
    \begin{tabular}{@{}lcccccc@{}}
        \toprule
        & \multicolumn{3}{c}{Train Set} & \multicolumn{3}{c}{Val. Set} \\
        \cmidrule(lr){2-4} \cmidrule(lr){5-7}
        Reg. & B16 & B32 & L14 & B16 & B32 & L14 \\
        \midrule
        $ Diag$ & .99 & .99 & .98 & .99 & .99 & .98 \\
$10^{-3}$ & .98 & .95 & .98 & .98 & .98 & .97 \\
$10^{-6}$ & .72 & .67 & .75 & .97 & .97 & .99 \\
$10^{-9}$ & .54 & .42 & .64 & .96 & .97 & .98 \\
\bottomrule
\end{tabular}
    }
\end{minipage}
\end{table}

\begin{figure}[!htb]
    \centering
    \begin{subfigure}{0.48\textwidth}
        \centering
        \includegraphics[width=\linewidth]{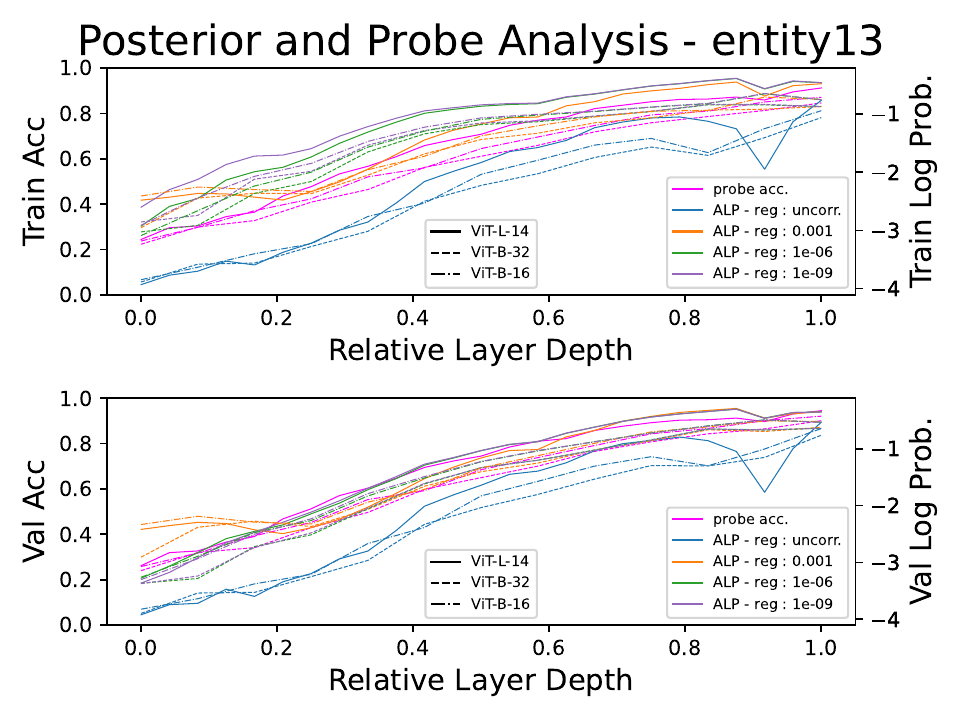}
        \caption{CLIP Model on entity13.}
        \label{fig:entity13-allnetallreg}
    \end{subfigure}
    \begin{subfigure}{0.48\textwidth}
        \centering
        \includegraphics[width=\linewidth]{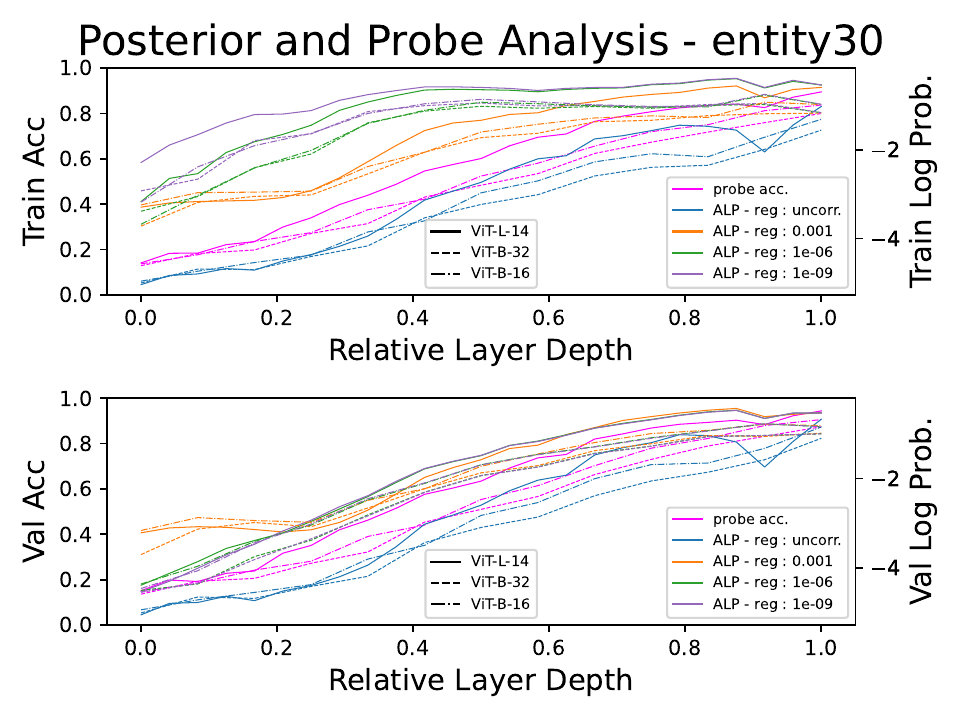}
        \caption{CLIP Model on entity30.}
        \label{fig:entity30-allnetallreg}
    \end{subfigure}
    \caption{Evolution of layerwise log posterior and linear probe accuracies for CLIP Models across varying regularization strengths, demonstrating correlations between log posterior and linear probe performance across the depth of various CLIP Models. Quantitative results are shown in \autoref{tab:breedsCLIPpearcorr1} and \autoref{tab:breedsCLIPpearcorr2}.}
    \label{fig:CLIPAnalysisEntity}
\end{figure}

\begin{figure}[!htb]
    \centering
    \begin{subfigure}{0.48\textwidth}
        \centering
        \includegraphics[width=\linewidth]{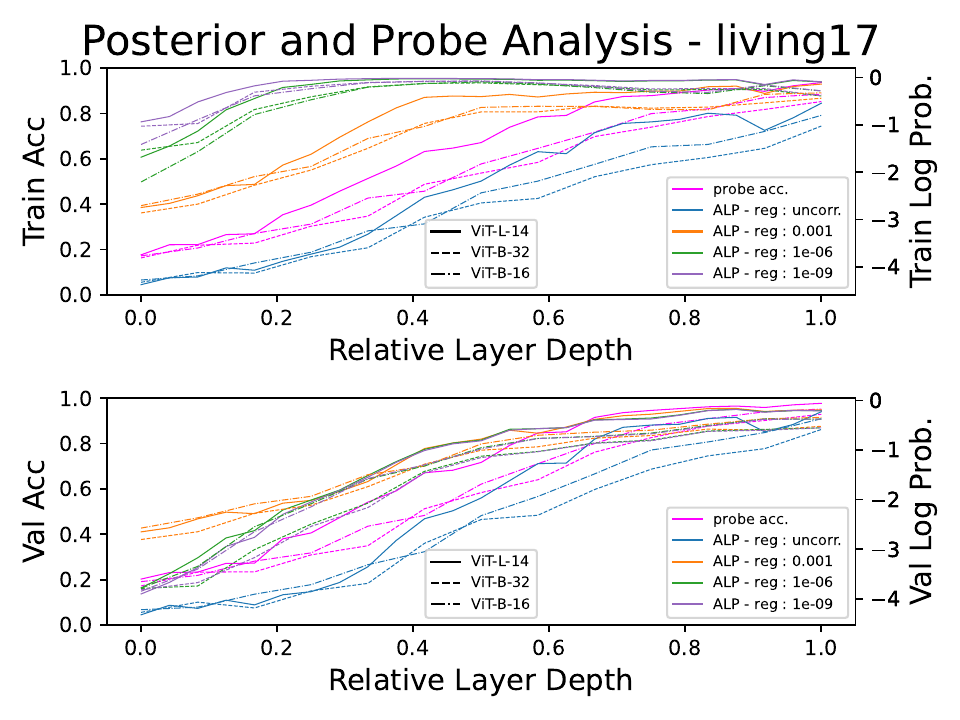}
        \caption{CLIP Models on living17.}
        \label{fig:living17-allnetallreg}
    \end{subfigure}
    \begin{subfigure}{0.48\textwidth}
        \centering
        \includegraphics[width=\linewidth]{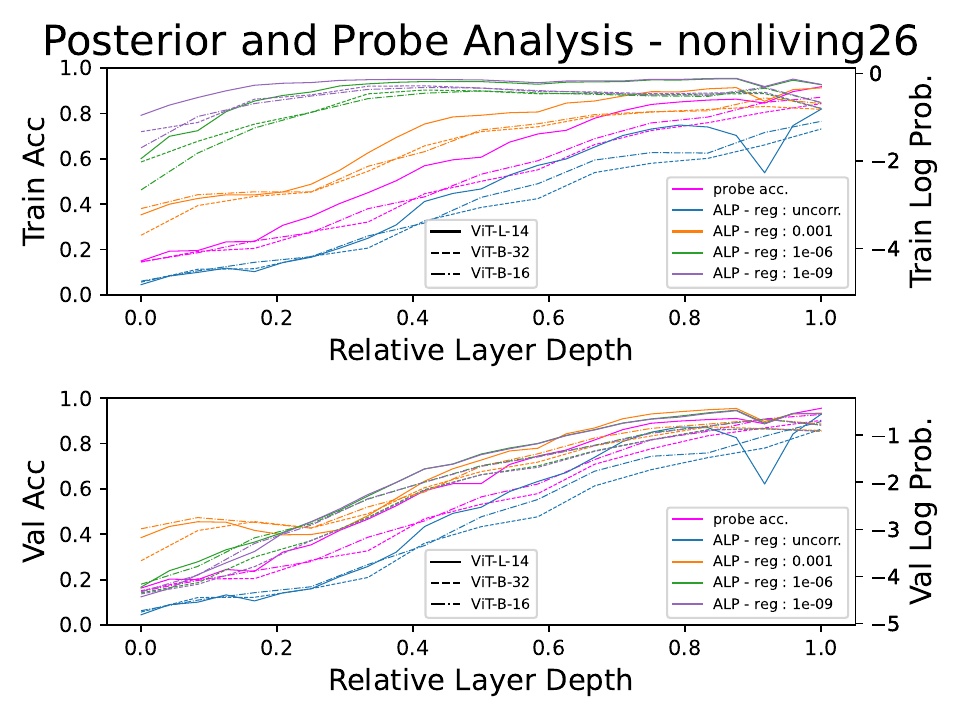}
        \caption{CLIP Models on nonliving26.}
        \label{fig:nonliving26-allnetallreg}
    \end{subfigure}
    \caption{Evolution of layerwise log posterior and linear probe accuracies for CLIP Models across     varying regularization strengths, demonstrating correlations between log posterior and linear probe performance across the depth of various CLIP Models. A more detailed breakdown of the results comparing the correlations on each dataset for various settings of regularizations corresponding to \autoref{fig:CLIPAnalysisEntity} and \autoref{fig:CLIPAnalysisLivingNonLiving} is shown in \autoref{fig:CLIPAnalysisBreeds-entity13-Part1} - \autoref{fig:CLIPAnalysisBreeds-nonliving26-Part2} of Appendix \ref{ssec:CLIPBreedsallNetPerRegPerData}.}
    \label{fig:CLIPAnalysisLivingNonLiving}
\end{figure}

For this experiment, we measured the correlation between average log posteriors and linear probe accuracy learned and computed over the Breeds training and validation set embeddings. The results are shown in \autoref{tab:breedsCLIPpearcorr1} and \autoref{tab:breedsCLIPpearcorr2} for respective datasets from \autoref{fig:CLIPAnalysisEntity} and \autoref{fig:CLIPAnalysisLivingNonLiving}. Based on those results, we demonstrate that average log posterior and linear probe performance correlates strongly across the depth of the network when measured via Pearson's correlation. This is across various settings of regularizations (both with Independence assumptions and Tikhonov\footnote{https://web.eecs.umich.edu/~aey/recent/regular.pdf} Regularization) of the class-wise covariance matrices for our learned average log posterior metric for various Breeds Datasets and CLIP Models.

Our results with a layerwise analysis of Pre-trained CLIP Models comparing our metric with a linear probe on internal activations help us assert that the log posterior is predictive of an embedding's downstream classification performance, even if a set of data that was used to generate the embedding wasn't part of the models training. With Pearson correlations greater than 0.9 for a majority of settings in \autoref{tab:breedsCLIPpearcorr1} and \autoref{tab:breedsCLIPpearcorr2}, we can confidently establish its value as a metric that can distinguish between good and bad-performing embeddings generated by various models. Additionally we also show the corresponding spearman correlations in \autoref{tab:EntityCLIPspearcorr} and \autoref{tab:LivingNonLivingCLIPspearcorr}. 



\begin{table*}[!htb]
\centering
\caption{Spearman's Correlation on Entity Datasets}
\label{tab:EntityCLIPspearcorr}
\begin{minipage}{.48\linewidth}
    \centering
    \caption*{Entity 13 Dataset}
    \setlength{\tabcolsep}{3pt}
    \scalebox{0.8}{
    \begin{tabular}{@{}lcccccc@{}}
        \toprule
        & \multicolumn{3}{c}{Train Set} & \multicolumn{3}{c}{Val. Set} \\
        \cmidrule(lr){2-4} \cmidrule(lr){5-7}
        Reg. & B16 & B32 & L14 & B16 & B32 & L14 \\
        \midrule
        $Diag$ & .98 & .99 & .95 & .99 & .99 & .94 \\
        $10^{-3}$ & .97 & .99 & .98 & .97 & .98 & .96 \\
        $10^{-6}$ & .99 & .97 & .99 & .99 & .99 & .99 \\
        $10^{-9}$ & .99 & .98 & .98 & .99 & .99 & .99 \\
        \bottomrule
    \end{tabular}
    }
\end{minipage}
\begin{minipage}{.48\linewidth}
    \centering
    \caption*{Entity 30 Dataset}
    \scalebox{0.8}{
    \begin{tabular}{@{}lcccccc@{}}
        \toprule
        & \multicolumn{3}{c}{Train Set} & \multicolumn{3}{c}{Val. Set} \\
        \cmidrule(lr){2-4} \cmidrule(lr){5-7}
        Reg. & B16 & B32 & L14 & B16 & B32 & L14 \\
        \midrule
        $Diag$ & .99 & .99 & .98 & 1.00 & .99 & .98 \\
        $10^{-3}$ & .98 & 1.0 & .99 & .97 & .98 & .96 \\
        $10^{-6}$ & .82 & .79 & .96 & .99 & 1.00 & .99 \\
        $10^{-9}$ & .72         & .67 & .91 & .99 & 1.00 & .99 \\
        \bottomrule
    \end{tabular}
    }
\end{minipage}
\end{table*}

\begin{table*}[!htb]
\centering
\caption{Spearman's Correlation on Living and Non Living Datasets}
\label{tab:LivingNonLivingCLIPspearcorr}
\begin{minipage}{.48\linewidth}
    \centering
    \caption*{Living 17 Dataset}
    \setlength{\tabcolsep}{3pt}
    \scalebox{0.8}{
    \begin{tabular}{@{}lcccccc@{}}
        \toprule
        & \multicolumn{3}{c}{Train Set} & \multicolumn{3}{c}{Val. Set} \\
        \cmidrule(lr){2-4} \cmidrule(lr){5-7}
        Reg. & B16 & B32 & L14 & B16 & B32 & L14 \\
        \midrule
        $Diag$ & 1.00 & .99 & .99 & 1.00 & .99 & .98 \\
        $10^{-3}$ & .9 & .97 & .98 & 1.00 & .99 & .99 \\
        $10^{-6}$ & .47 & .41 & .42 & .99 & 1.00 & .99 \\
        $10^{-9}$ & .3 & .16 & .26 & 1.00 & 1.00 & .99 \\
        \bottomrule
    \end{tabular}
    }
\end{minipage}%
\begin{minipage}{.48\linewidth}
    \centering
    \caption*{Non Living 26 Dataset}
    \scalebox{0.8}{
    \begin{tabular}{@{}lcccccc@{}}
        \toprule
        & \multicolumn{3}{c}{Train Set} & \multicolumn         {3}{c}{Val. Set} \\
        \cmidrule(lr){2-4} \cmidrule(lr){5-7}
        Reg. & B16 & B32 & L14 & B16 & B32 & L14 \\
        \midrule
        $Diag$ & .99 & 1.00 & .97 & 1.00 & 1.00 & .97 \\
        $10^{-3}$ & .99 & .99 & .98 & .97 & .97 & .95 \\
        $10^{-6}$ & .62 & .48 & .77 & .99 & .98 & .99 \\
        $10^{-9}$ & .46 & .26 & .61 & .99 & .98 & .99 \\
        \bottomrule
    \end{tabular}
    }
\end{minipage}
\end{table*}

\section{Conclusion}
This work introduces a novel method for evaluating pre-trained models. Instead of using costly and time-consuming fine-tuned downstream tasks for evaluation, we propose using the consistency between entity embeddings and their associated meta-features as a performance metric. Our method has been effectively tested across various domains and datasets in relational datasets, Natural Language Processing, and Computer Vision, providing a more efficient and equally rigorous alternative for pre-trained model evaluation.

\newpage

\bibliography{iclr2024_conference}

\begin{thebibliography}{27}
\providecommand{\natexlab}[1]{#1}
\providecommand{\url}[1]{\texttt{#1}}
\expandafter\ifx\csname urlstyle\endcsname\relax
  \providecommand{\doi}[1]{doi: #1}\else
  \providecommand{\doi}{doi: \begingroup \urlstyle{rm}\Url}\fi

\bibitem[Ahuja et~al.(2023)Ahuja, Diddee, Hada, Ochieng, Ramesh, Jain, Nambi, Ganu, Segal, Ahmed, Bali, and Sitaram]{ahuja2023mega}
Kabir Ahuja, Harshita Diddee, Rishav Hada, Millicent Ochieng, Krithika Ramesh, Prachi Jain, Akshay Nambi, Tanuja Ganu, Sameer Segal, Mohamed Ahmed, Kalika Bali, and Sunayana Sitaram.
\newblock {MEGA}: Multilingual evaluation of generative {AI}.
\newblock In \emph{The 2023 Conference on Empirical Methods in Natural Language Processing}, 2023.
\newblock URL \url{https://openreview.net/forum?id=jmopGajkFY}.

\bibitem[Bang et~al.(2023)Bang, Cahyawijaya, Lee, Dai, Su, Wilie, Lovenia, Ji, Yu, Chung, Do, Xu, and Fung]{bang-etal-2023-multitask}
Yejin Bang, Samuel Cahyawijaya, Nayeon Lee, Wenliang Dai, Dan Su, Bryan Wilie, Holy Lovenia, Ziwei Ji, Tiezheng Yu, Willy Chung, Quyet~V. Do, Yan Xu, and Pascale Fung.
\newblock A multitask, multilingual, multimodal evaluation of {C}hat{GPT} on reasoning, hallucination, and interactivity.
\newblock In Jong~C. Park, Yuki Arase, Baotian Hu, Wei Lu, Derry Wijaya, Ayu Purwarianti, and Adila~Alfa Krisnadhi (eds.), \emph{Proceedings of the 13th International Joint Conference on Natural Language Processing and the 3rd Conference of the Asia-Pacific Chapter of the Association for Computational Linguistics (Volume 1: Long Papers)}, pp.\  675--718, Nusa Dua, Bali, November 2023. Association for Computational Linguistics.
\newblock URL \url{https://aclanthology.org/2023.ijcnlp-main.45}.

\bibitem[Breiman(2001)]{breiman2001random}
Leo Breiman.
\newblock Random forests.
\newblock \emph{Machine learning}, 45:\penalty0 5--32, 2001.

\bibitem[Brown et~al.(2020)Brown, Mann, Ryder, Subbiah, Kaplan, Dhariwal, Neelakantan, Shyam, Sastry, Askell, et~al.]{brown2020language}
Tom Brown, Benjamin Mann, Nick Ryder, Melanie Subbiah, Jared~D Kaplan, Prafulla Dhariwal, Arvind Neelakantan, Pranav Shyam, Girish Sastry, Amanda Askell, et~al.
\newblock Language models are few-shot learners.
\newblock \emph{Advances in neural information processing systems}, 33:\penalty0 1877--1901, 2020.

\bibitem[Chen et~al.(2019)Chen, Stanovsky, Singh, and Gardner]{chen-etal-2019-evaluating}
Anthony Chen, Gabriel Stanovsky, Sameer Singh, and Matt Gardner.
\newblock Evaluating question answering evaluation.
\newblock In Adam Fisch, Alon Talmor, Robin Jia, Minjoon Seo, Eunsol Choi, and Danqi Chen (eds.), \emph{Proceedings of the 2nd Workshop on Machine Reading for Question Answering}, pp.\  119--124, Hong Kong, China, November 2019. Association for Computational Linguistics.
\newblock \doi{10.18653/v1/D19-5817}.
\newblock URL \url{https://aclanthology.org/D19-5817}.

\bibitem[Dai et~al.(2023)Dai, Fan, Zhuang, Jain, Yeh, Wang, Wang, Zheng, and Zhang]{dai2023pdt}
Xin Dai, Yujie Fan, Zhongfang Zhuang, Shubham Jain, Chin-Chia~Michael Yeh, Junpeng Wang, Liang Wang, Yan Zheng, and Wei Zhang.
\newblock Pdt: Pretrained dual transformers for time-aware bipartite graphs.
\newblock \emph{arXiv preprint arXiv:2306.01913}, 2023.

\bibitem[Deng et~al.(2009)Deng, Dong, Socher, Li, Li, and Fei-Fei]{5206848}
Jia Deng, Wei Dong, Richard Socher, Li-Jia Li, Kai Li, and Li~Fei-Fei.
\newblock Imagenet: A large-scale hierarchical image database.
\newblock In \emph{2009 IEEE Conference on Computer Vision and Pattern Recognition}, pp.\  248--255, 2009.
\newblock \doi{10.1109/CVPR.2009.5206848}.

\bibitem[Engstrom et~al.(2019{\natexlab{a}})Engstrom, Ilyas, Santurkar, and Tsipras]{robustness}
Logan Engstrom, Andrew Ilyas, Shibani Santurkar, and Dimitris Tsipras.
\newblock Robustness (python library), 2019{\natexlab{a}}.
\newblock URL \url{https://github.com/MadryLab/robustness}.

\bibitem[Engstrom et~al.(2019{\natexlab{b}})Engstrom, Ilyas, Santurkar, Tsipras, Tran, and Madry]{engstrom2019adversarial}
Logan Engstrom, Andrew Ilyas, Shibani Santurkar, Dimitris Tsipras, Brandon Tran, and Aleksander Madry.
\newblock Adversarial robustness as a prior for learned representations, 2019{\natexlab{b}}.

\bibitem[Gurnee \& Tegmark(2023)Gurnee and Tegmark]{gurnee2023language}
Wes Gurnee and Max Tegmark.
\newblock Language models represent space and time.
\newblock \emph{arXiv preprint arXiv:2310.02207}, 2023.

\bibitem[Harper \& Konstan(2015)Harper and Konstan]{harper2015movielens}
F~Maxwell Harper and Joseph~A Konstan.
\newblock The movielens datasets: History and context.
\newblock \emph{Acm transactions on interactive intelligent systems (tiis)}, 5\penalty0 (4):\penalty0 1--19, 2015.

\bibitem[Kang \& McAuley(2018)Kang and McAuley]{SASRec}
Wang-Cheng Kang and Julian McAuley.
\newblock Self-attentive sequential recommendation.
\newblock In \emph{2018 IEEE international conference on data mining (ICDM)}, pp.\  197--206. IEEE, 2018.

\bibitem[Kenton \& Toutanova(2019)Kenton and Toutanova]{kenton2019bert}
Jacob Devlin Ming-Wei~Chang Kenton and Lee~Kristina Toutanova.
\newblock Bert: Pre-training of deep bidirectional transformers for language understanding.
\newblock In \emph{Proceedings of naacL-HLT}, volume~1, pp.\ ~2, 2019.

\bibitem[Kusupati et~al.(2022)Kusupati, Bhatt, Rege, Wallingford, Sinha, Ramanujan, Howard-Snyder, Chen, Kakade, Jain, and Farhadi]{Kusupati2022MatryoshkaRL}
Aditya Kusupati, Gantavya Bhatt, Aniket Rege, Matthew Wallingford, Aditya Sinha, Vivek Ramanujan, William Howard-Snyder, Kaifeng Chen, Sham~M. Kakade, Prateek Jain, and Ali Farhadi.
\newblock Matryoshka representation learning.
\newblock In \emph{Neural Information Processing Systems}, 2022.
\newblock URL \url{https://api.semanticscholar.org/CorpusID:252683450}.

\bibitem[Liang et~al.(2023)Liang, Bommasani, Lee, Tsipras, Soylu, Yasunaga, Zhang, Narayanan, Wu, Kumar, Newman, Yuan, Yan, Zhang, Cosgrove, Manning, Re, Acosta-Navas, Hudson, Zelikman, Durmus, Ladhak, Rong, Ren, Yao, WANG, Santhanam, Orr, Zheng, Yuksekgonul, Suzgun, Kim, Guha, Chatterji, Khattab, Henderson, Huang, Chi, Xie, Santurkar, Ganguli, Hashimoto, Icard, Zhang, Chaudhary, Wang, Li, Mai, Zhang, and Koreeda]{liang2023holistic}
Percy Liang, Rishi Bommasani, Tony Lee, Dimitris Tsipras, Dilara Soylu, Michihiro Yasunaga, Yian Zhang, Deepak Narayanan, Yuhuai Wu, Ananya Kumar, Benjamin Newman, Binhang Yuan, Bobby Yan, Ce~Zhang, Christian~Alexander Cosgrove, Christopher~D Manning, Christopher Re, Diana Acosta-Navas, Drew~Arad Hudson, Eric Zelikman, Esin Durmus, Faisal Ladhak, Frieda Rong, Hongyu Ren, Huaxiu Yao, Jue WANG, Keshav Santhanam, Laurel Orr, Lucia Zheng, Mert Yuksekgonul, Mirac Suzgun, Nathan Kim, Neel Guha, Niladri~S. Chatterji, Omar Khattab, Peter Henderson, Qian Huang, Ryan~Andrew Chi, Sang~Michael Xie, Shibani Santurkar, Surya Ganguli, Tatsunori Hashimoto, Thomas Icard, Tianyi Zhang, Vishrav Chaudhary, William Wang, Xuechen Li, Yifan Mai, Yuhui Zhang, and Yuta Koreeda.
\newblock Holistic evaluation of language models.
\newblock \emph{Transactions on Machine Learning Research}, 2023.
\newblock ISSN 2835-8856.
\newblock URL \url{https://openreview.net/forum?id=iO4LZibEqW}.
\newblock Featured Certification, Expert Certification.

\bibitem[Mikolov et~al.(2013)Mikolov, Chen, Corrado, and Dean]{Mikolov2013EfficientEO}
Tomas Mikolov, Kai Chen, Gregory~S. Corrado, and Jeffrey Dean.
\newblock Efficient estimation of word representations in vector space.
\newblock In \emph{International Conference on Learning Representations}, 2013.
\newblock URL \url{https://api.semanticscholar.org/CorpusID:5959482}.

\bibitem[Miller(1994)]{miller-1994-wordnet}
George~A. Miller.
\newblock {W}ord{N}et: A lexical database for {E}nglish.
\newblock In \emph{{H}uman {L}anguage {T}echnology: Proceedings of a Workshop held at {P}lainsboro, {N}ew {J}ersey, {M}arch 8-11, 1994}, 1994.
\newblock URL \url{https://aclanthology.org/H94-1111}.

\bibitem[OpenAI(2023)]{openai2023gpt4}
OpenAI.
\newblock Gpt-4 technical report, 2023.

\bibitem[Radford et~al.(2021)Radford, Kim, Hallacy, Ramesh, Goh, Agarwal, Sastry, Askell, Mishkin, Clark, Krueger, and Sutskever]{radford2021learning}
Alec Radford, Jong~Wook Kim, Chris Hallacy, Aditya Ramesh, Gabriel Goh, Sandhini Agarwal, Girish Sastry, Amanda Askell, Pamela Mishkin, Jack Clark, Gretchen Krueger, and Ilya Sutskever.
\newblock Learning transferable visual models from natural language supervision, 2021.

\bibitem[Raffel et~al.(2020)Raffel, Shazeer, Roberts, Lee, Narang, Matena, Zhou, Li, and Liu]{raffel2020exploring}
Colin Raffel, Noam Shazeer, Adam Roberts, Katherine Lee, Sharan Narang, Michael Matena, Yanqi Zhou, Wei Li, and Peter~J Liu.
\newblock Exploring the limits of transfer learning with a unified text-to-text transformer.
\newblock \emph{The Journal of Machine Learning Research}, 21\penalty0 (1):\penalty0 5485--5551, 2020.

\bibitem[Santurkar et~al.(2019)Santurkar, Tsipras, Tran, Ilyas, Engstrom, and Madry]{santurkar2019image}
Shibani Santurkar, Dimitris Tsipras, Brandon Tran, Andrew Ilyas, Logan Engstrom, and Aleksander Madry.
\newblock Image synthesis with a single (robust) classifier, 2019.

\bibitem[Santurkar et~al.(2020)Santurkar, Tsipras, and Madry]{santurkar2020breeds}
Shibani Santurkar, Dimitris Tsipras, and Aleksander Madry.
\newblock Breeds: Benchmarks for subpopulation shift, 2020.

\bibitem[Touvron et~al.(2023)Touvron, Martin, Stone, Albert, Almahairi, Babaei, Bashlykov, Batra, Bhargava, Bhosale, et~al.]{touvron2023llama}
Hugo Touvron, Louis Martin, Kevin Stone, Peter Albert, Amjad Almahairi, Yasmine Babaei, Nikolay Bashlykov, Soumya Batra, Prajjwal Bhargava, Shruti Bhosale, et~al.
\newblock Llama 2: Open foundation and fine-tuned chat models.
\newblock \emph{arXiv preprint arXiv:2307.09288}, 2023.

\bibitem[Wang et~al.(2023)Wang, Lyu, Ji, Zhang, Yu, Shi, and Tu]{wang-etal-2023-document-level}
Longyue Wang, Chenyang Lyu, Tianbo Ji, Zhirui Zhang, Dian Yu, Shuming Shi, and Zhaopeng Tu.
\newblock Document-level machine translation with large language models.
\newblock In Houda Bouamor, Juan Pino, and Kalika Bali (eds.), \emph{Proceedings of the 2023 Conference on Empirical Methods in Natural Language Processing}, pp.\  16646--16661, Singapore, December 2023. Association for Computational Linguistics.
\newblock \doi{10.18653/v1/2023.emnlp-main.1036}.
\newblock URL \url{https://aclanthology.org/2023.emnlp-main.1036}.

\bibitem[Workshop et~al.(2022)Workshop, Scao, Fan, Akiki, Pavlick, Ili{\'c}, Hesslow, Castagn{\'e}, Luccioni, Yvon, et~al.]{workshop2022bloom}
BigScience Workshop, Teven~Le Scao, Angela Fan, Christopher Akiki, Ellie Pavlick, Suzana Ili{\'c}, Daniel Hesslow, Roman Castagn{\'e}, Alexandra~Sasha Luccioni, Fran{\c{c}}ois Yvon, et~al.
\newblock Bloom: A 176b-parameter open-access multilingual language model.
\newblock \emph{arXiv preprint arXiv:2211.05100}, 2022.

\bibitem[Yang et~al.(2019)Yang, Dai, Yang, Carbonell, Salakhutdinov, and Le]{yang2019xlnet}
Zhilin Yang, Zihang Dai, Yiming Yang, Jaime Carbonell, Russ~R Salakhutdinov, and Quoc~V Le.
\newblock Xlnet: Generalized autoregressive pretraining for language understanding.
\newblock \emph{Advances in neural information processing systems}, 32, 2019.

\bibitem[Zheng et~al.(2023)Zheng, Wang, Yeh, Fan, Chen, Wang, and Zhang]{zheng2023embeddingtree}
Yan Zheng, Junpeng Wang, Chin-Chia~Michael Yeh, Yujie Fan, Huiyuan Chen, Liang Wang, and Wei Zhang.
\newblock Embeddingtree: Hierarchical exploration of entity features in embedding.
\newblock In \emph{2023 IEEE 16th Pacific Visualization Symposium (PacificVis)}, pp.\  217--221. IEEE, 2023.

\end{thebibliography}
\bibliographystyle{iclr2024_conference}

\appendix
\section{Appendix}
\label{sec:appendix}

\subsection{Fixed regularization cross Model analysis of CLIP Embeddings on the BREEDS hierarchy}
\label{ssec:CLIPBreedsallNetPerRegPerData}
\begin{figure}[H]
    \centering
    \begin{subfigure}{0.48\textwidth}
        \centering
        \includegraphics[width=\linewidth]{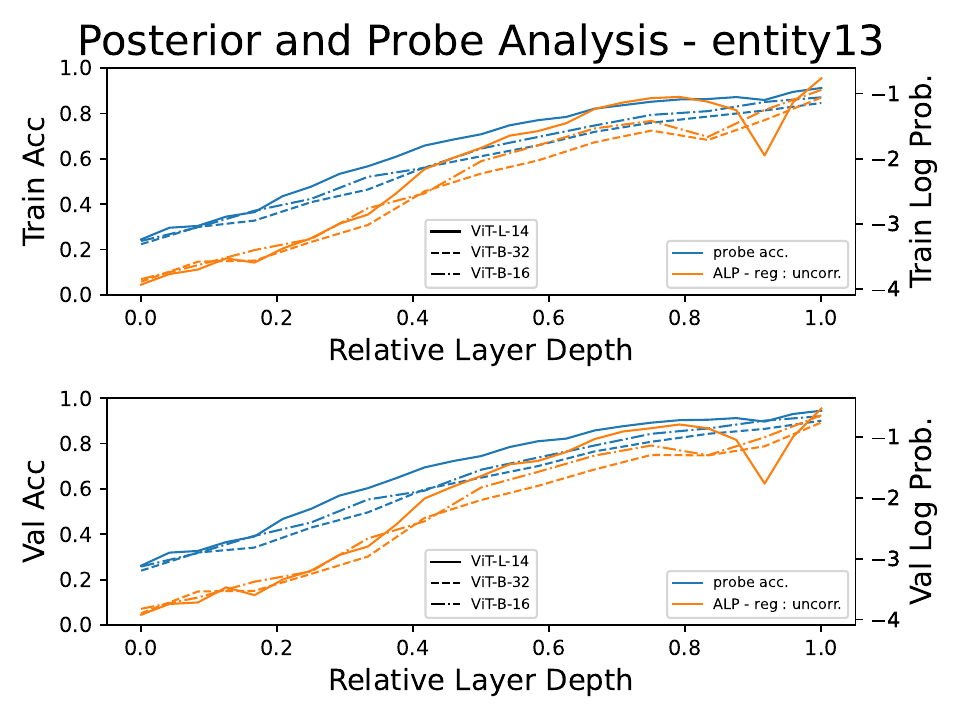}
        \caption{Regularization - Uncorr.}
        \label{fig:uncorr}
    \end{subfigure}
    \begin{subfigure}{0.48\textwidth}
        \centering
        \includegraphics[width=\linewidth]{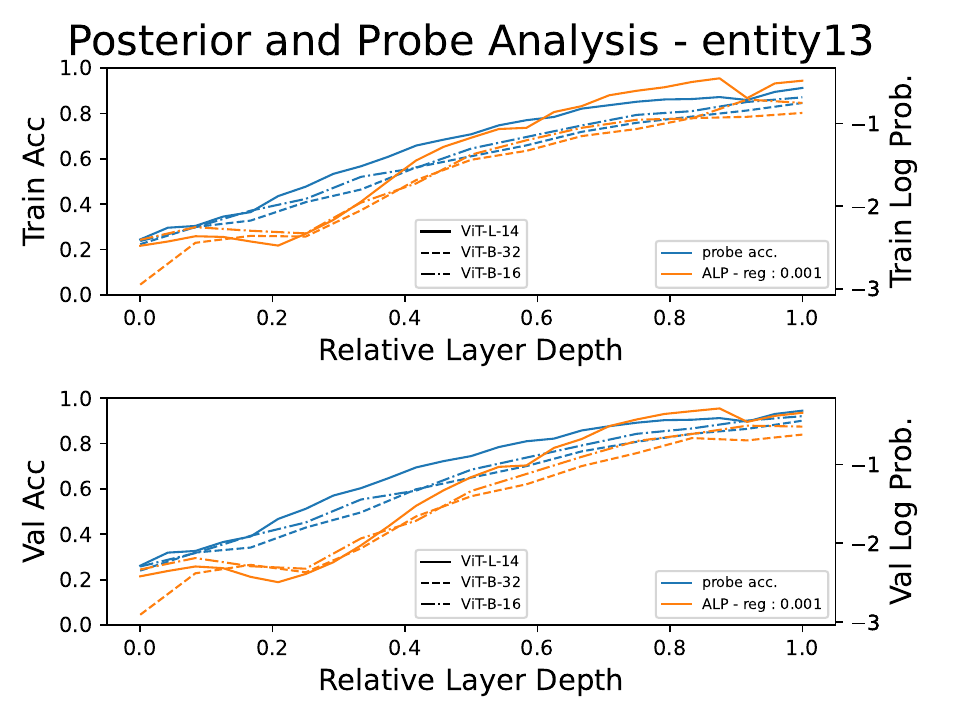}
        \caption{Regularization  $10^{-3}$.}
        \label{fig:reg1e-3}
    \end{subfigure}
    \caption{Regularization-wise analysis of layerwise ALP vs. linear probe performance for entity 13 breeds dataset.}
    \label{fig:CLIPAnalysisBreeds-entity13-Part1}
\end{figure}

\begin{figure}[H]
    \centering
    \begin{subfigure}{0.48\textwidth}
        \centering
        \includegraphics[width=\linewidth]{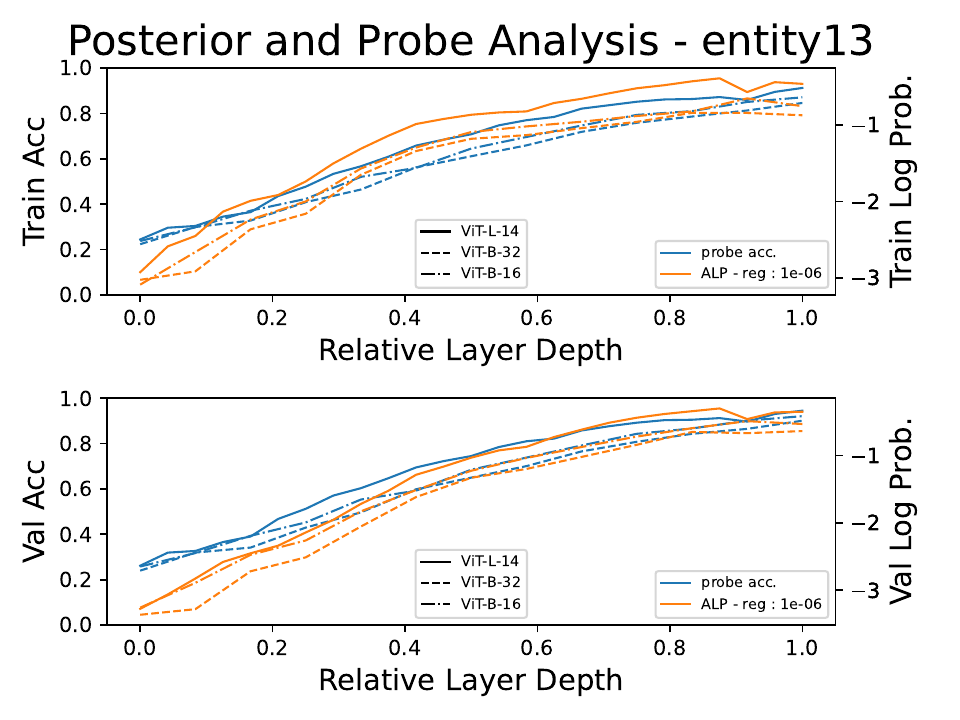}
        \caption{Regularization $10^{-6}$.}
        \label{fig:reg1e-6}
    \end{subfigure}
    \begin{subfigure}{0.48\textwidth}
        \centering
        \includegraphics[width=\linewidth]{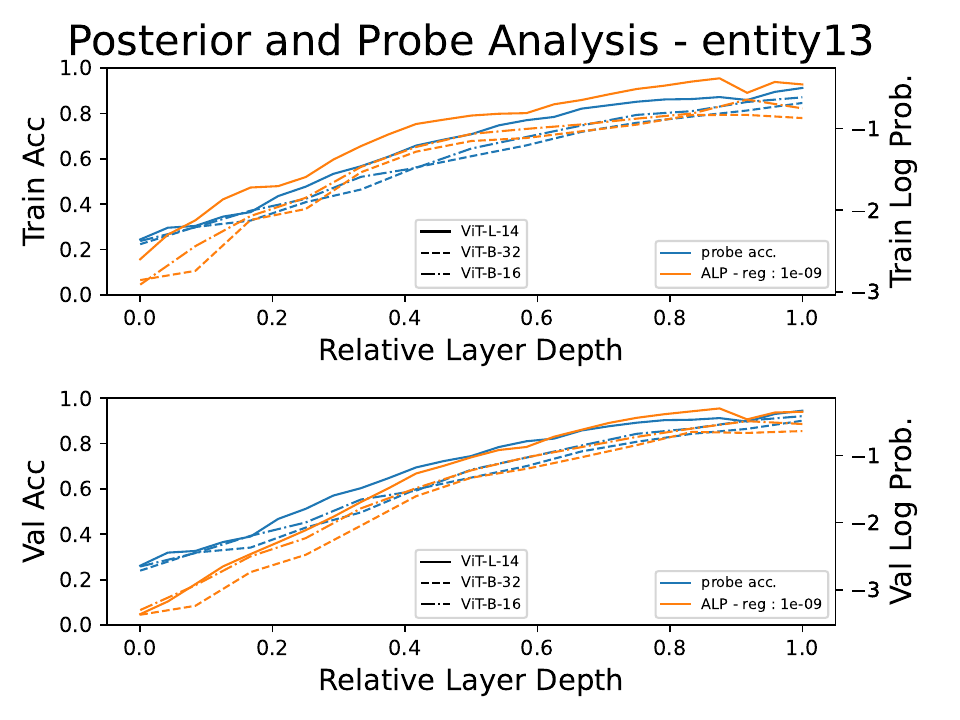}
        \caption{Regularization $10^{-9}$.}
        \label{fig:reg1e-9}
    \end{subfigure}
    \caption{Regularization-wise analysis of layerwise ALP vs. linear probe performance for entity 13 breeds dataset.}
    \label{fig:CLIPAnalysisBreeds-entity13-Part2}
\end{figure}

\begin{figure}[bp]
    \centering
    \begin{subfigure}{0.48\textwidth}
        \centering
        \includegraphics[width=\linewidth]{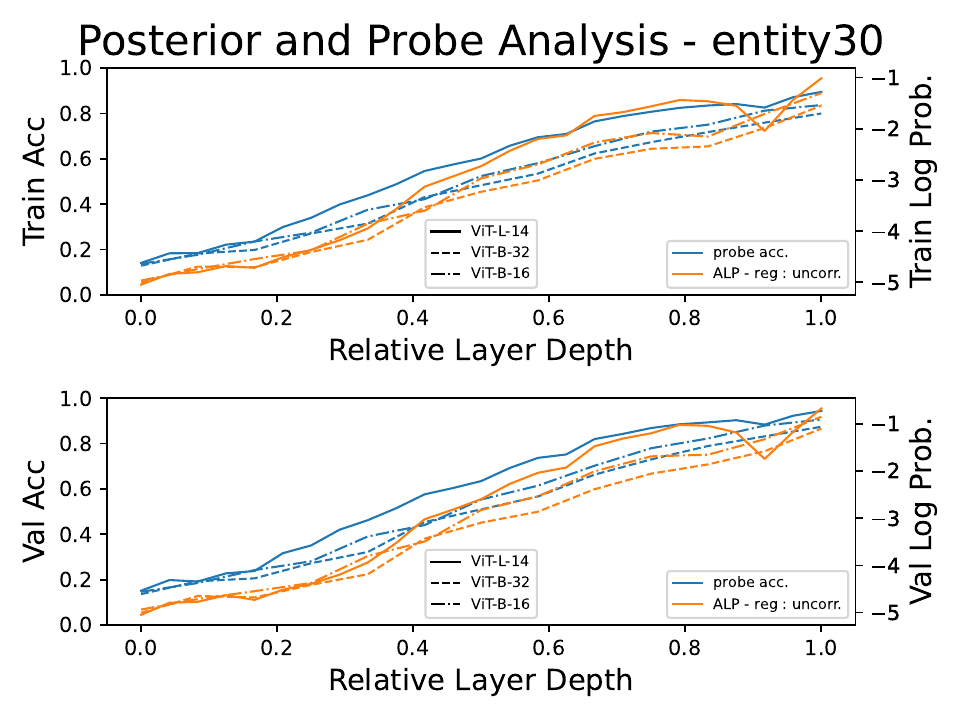}
        \caption{Regularization - Uncorr.}
        \label{fig:CLIPposteriorAnalysisBreeds-allnetperrege30_diag}
    \end{subfigure}
    \begin{subfigure}{0.48\textwidth}
        \centering
        \includegraphics[width=\linewidth]{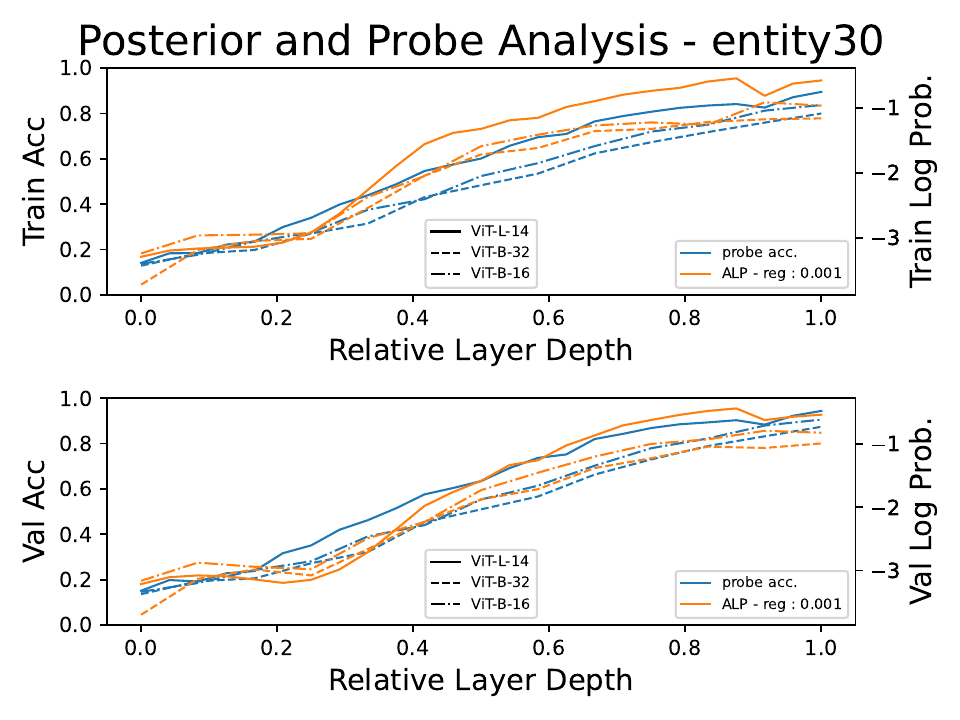}
        \caption{Regularization $10^{-3}$.}
        \label{fig:CLIPposteriorAnalysisBreeds-allnetperrege30_1e-3_demo}
    \end{subfigure}
    \caption{Regularization wise analysis of layerwise ALP vs. linear probe performance for entity 30 breeds dataset.}
    \label{fig:CLIPAnalysisBreeds-entity30-Part1}
\end{figure}

\begin{figure}[bp]
    \centering
    \begin{subfigure}{0.48\textwidth}
        \centering
        \includegraphics[width=\linewidth]{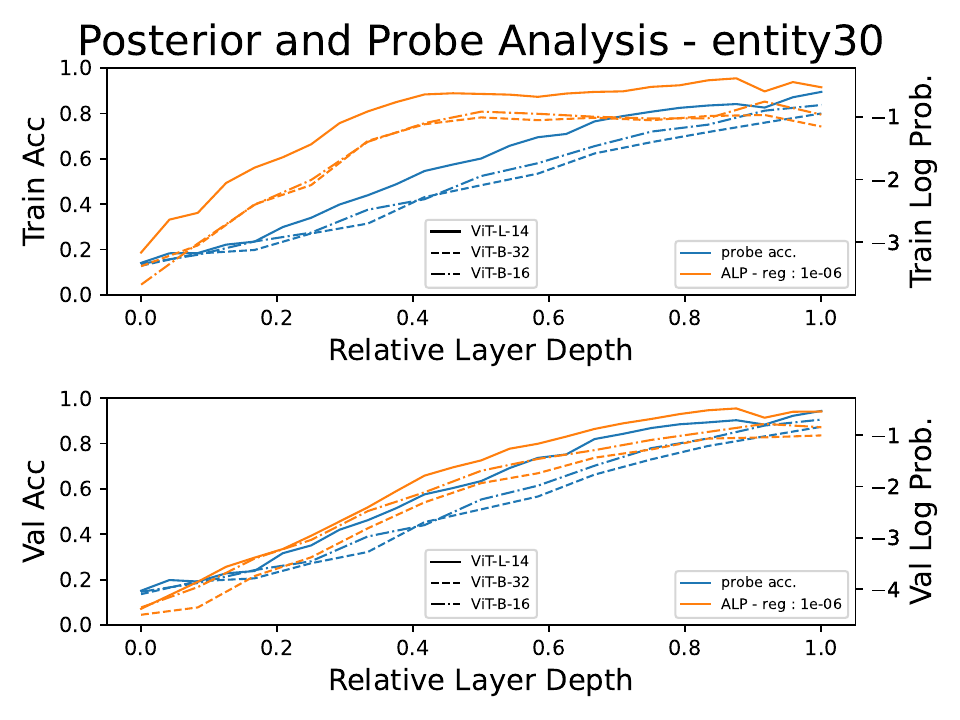}
        \caption{Regularization $10^{-6}$.}
        \label{fig:CLIPposteriorAnalysisBreeds-allnetperrege30_1e-6_demo}
    \end{subfigure}
    \begin{subfigure}{0.48\textwidth}
        \centering
        \includegraphics[width=\linewidth]{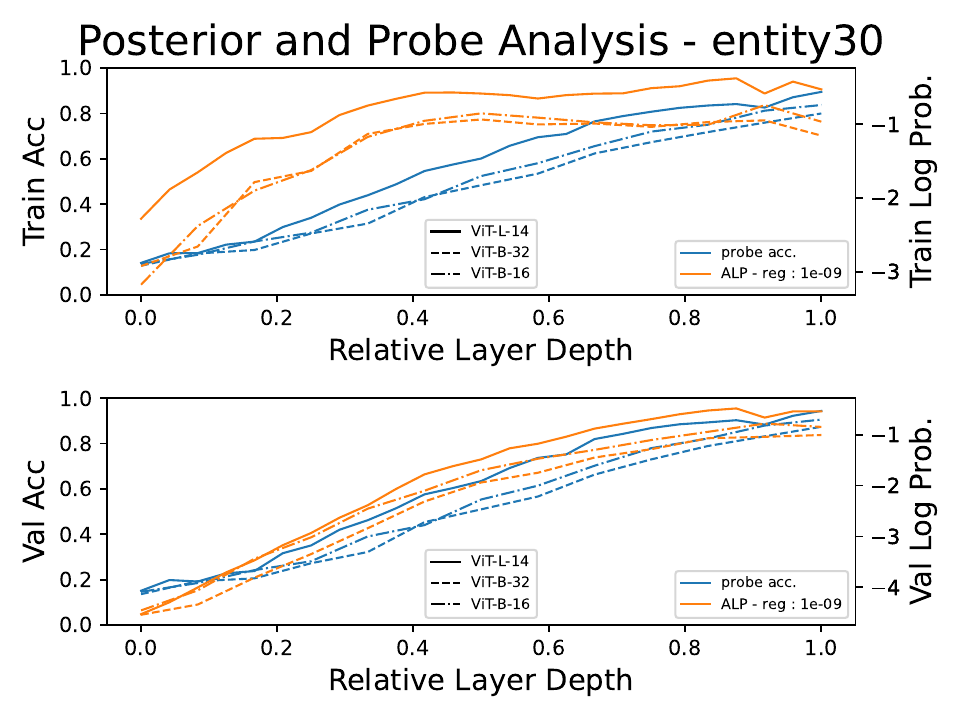}
        \caption{Regularization $10^{-9}$.}
        \label{fig:CLIPposteriorAnalysisBreeds-allnetperrege30_1e-9_demo}
    \end{subfigure}
    \caption{Regularization wise analysis of layerwise ALP vs. linear probe performance for entity 30 breeds dataset.}
    \label{fig:CLIPAnalysisBreeds-entity30-Part2}
\end{figure}

\begin{figure}[bp]
    \centering
    \begin{subfigure}{0.48\textwidth}
        \centering
        \includegraphics[width=\linewidth]{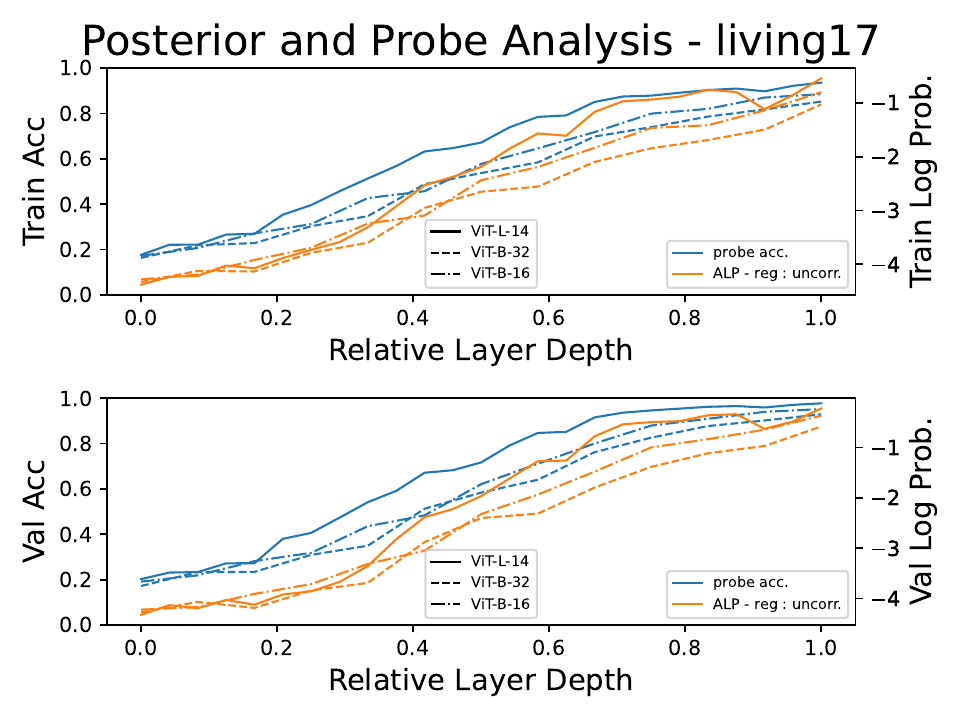}
        \caption{Regularization - Uncorr.}
        \label{fig:CLIPposteriorAnalysisBreeds-allnetperregl17_diag}
    \end{subfigure}
    \begin{subfigure}{0.48\textwidth}
        \centering
        \includegraphics[width=\linewidth]{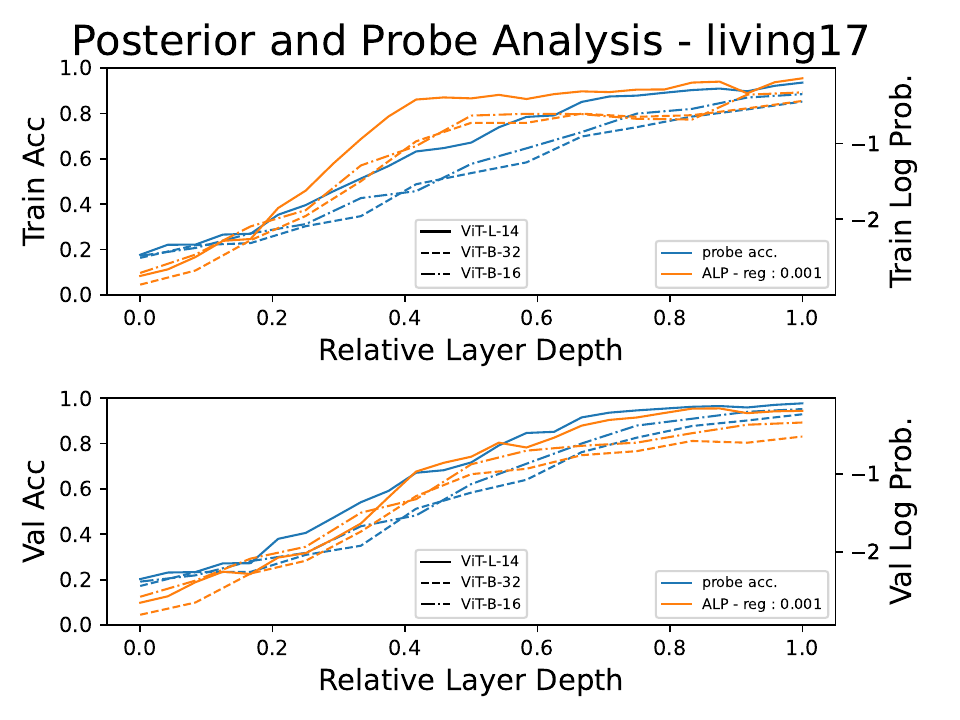}
        \caption{Regularization $10^{-3}$.}
        \label{fig:CLIPposteriorAnalysisBreeds-allnetperregl17_1e-3_demo}
    \end{subfigure}
    \caption{Regularization wise analysis of layerwise ALP vs. linear probe performance for living 17 breeds dataset.}
    \label{fig:CLIPAnalysisBreeds-living17-Part1}
\end{figure}

\begin{figure}[bp]
    \centering
    \begin{subfigure}{0.48\textwidth}
        \centering
        \includegraphics[width=\linewidth]{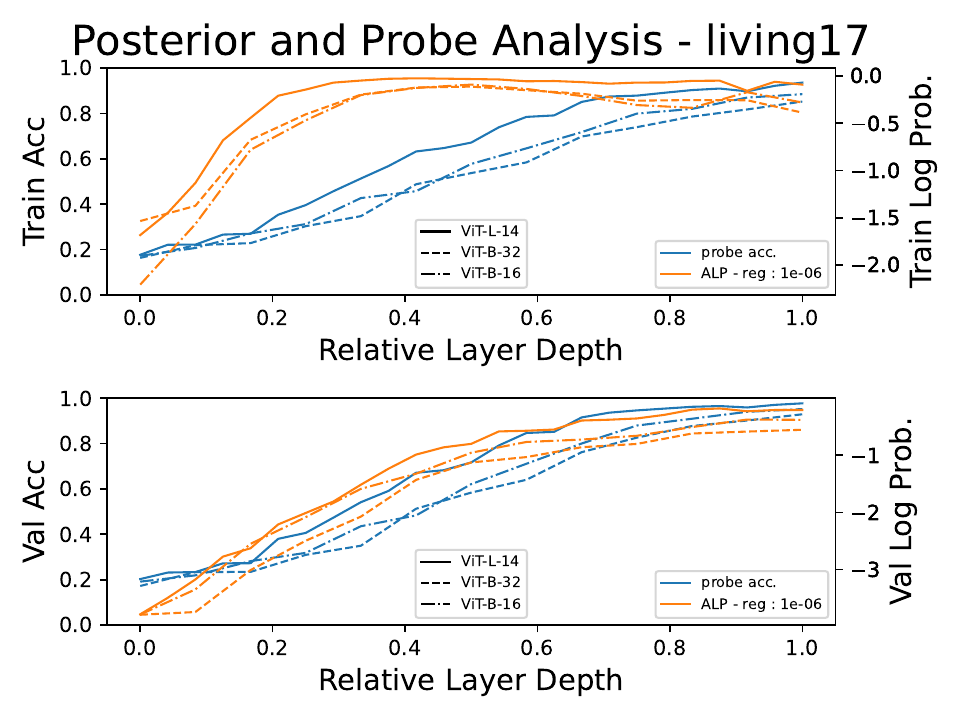}
        \caption{Regularization $10^{-6}$.}
        \label{fig:CLIPposteriorAnalysisBreeds-allnetperregl17_1e-6_demo}
    \end{subfigure}
    \begin{subfigure}{0.48\textwidth}
        \centering
        \includegraphics[width=\linewidth]{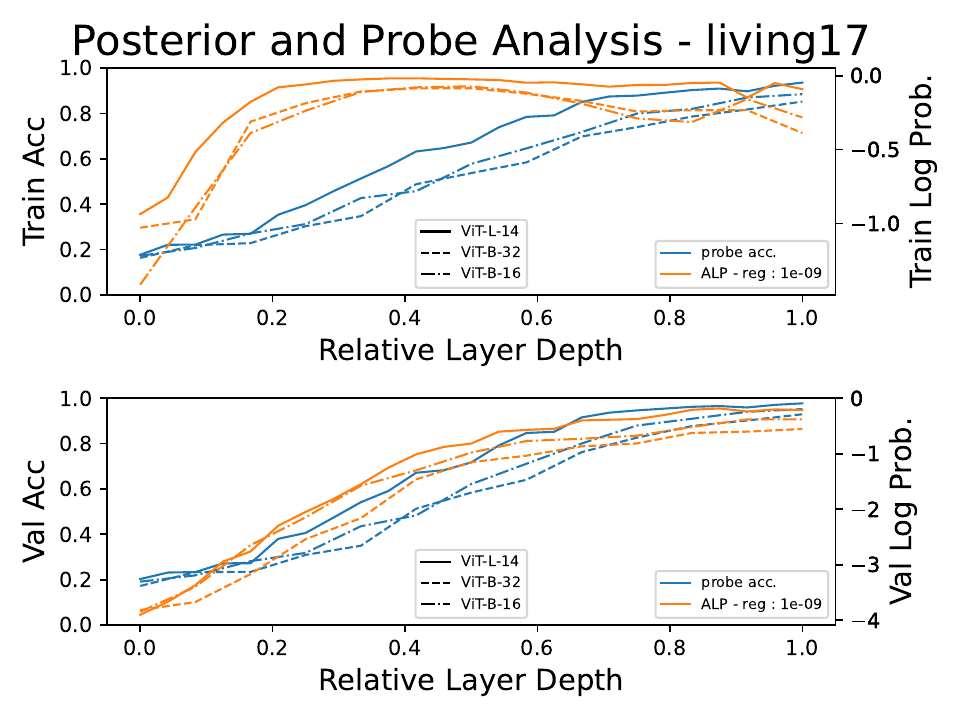}
        \caption{Regularization $10^{-9}$.}
        \label{fig:CLIPposteriorAnalysisBreeds-allnetperregl17_1e-9_demo}
    \end{subfigure}
    \caption{Regularization wise analysis of layerwise ALP vs. linear probe performance for living 17 breeds dataset.}
    \label{fig:CLIPAnalysisBreeds-living17-Part2}
\end{figure}

\begin{figure}[bp]
    \centering
    \begin{subfigure}{0.48\textwidth}
        \centering
        \includegraphics[width=\linewidth]{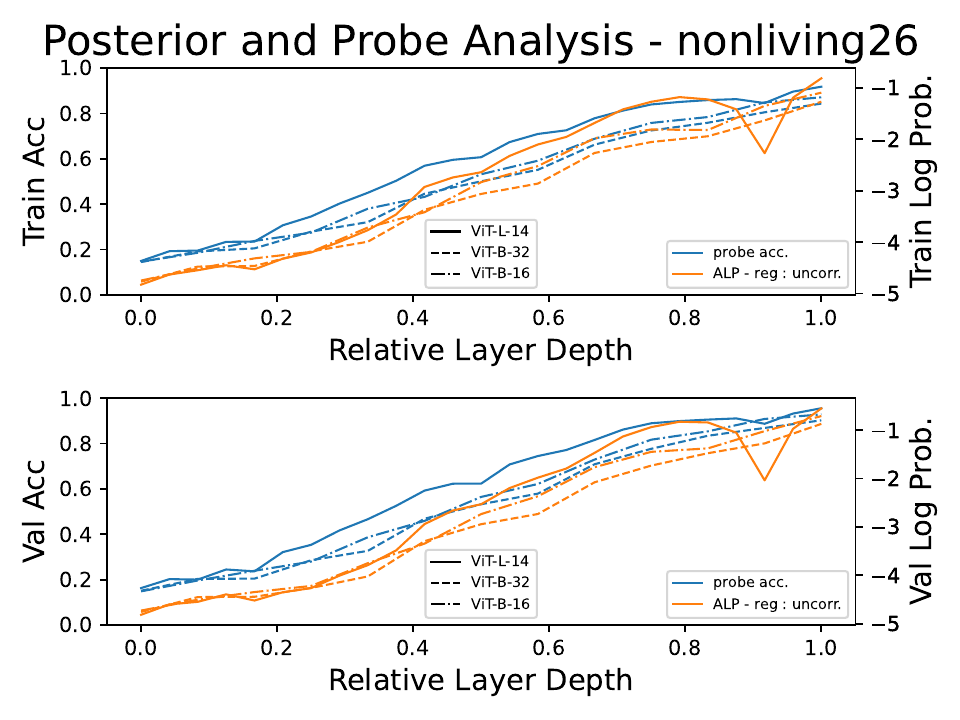}
        \caption{Regularization - Uncorr.}
        \label{fig:CLIPposteriorAnalysisBreeds-allnetperregnl26_diag}
    \end{subfigure}
    \begin{subfigure}{0.48\textwidth}
        \centering
        \includegraphics[width=\linewidth]{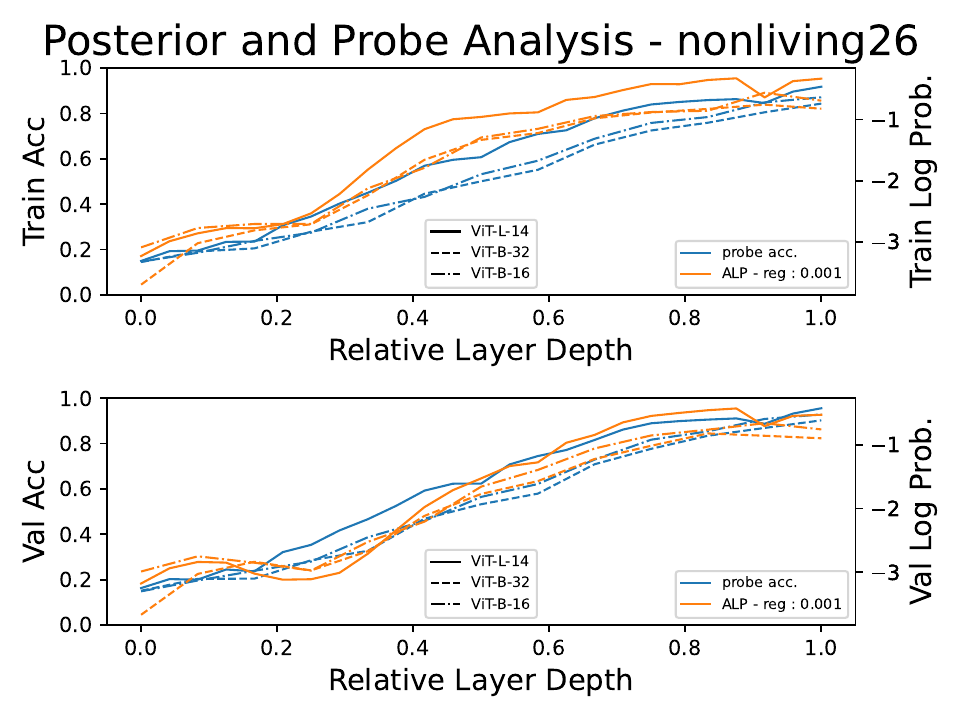}
        \caption{Regularization $10^{-3}$.}
        \label{fig:CLIPposteriorAnalysisBreeds-allnetperregnl26_1e-3_demo}
    \end{subfigure}
    \caption{Regularization-wise analysis of layerwise ALP vs. linear probe performance for nonliving 26 breeds dataset.}
    \label{fig:CLIPAnalysisBreeds-nonliving26-Part1}
\end{figure}

\begin{figure}[bp]
    \centering
    \begin{subfigure}{0.48\textwidth}
        \centering
        \includegraphics[width=\linewidth]{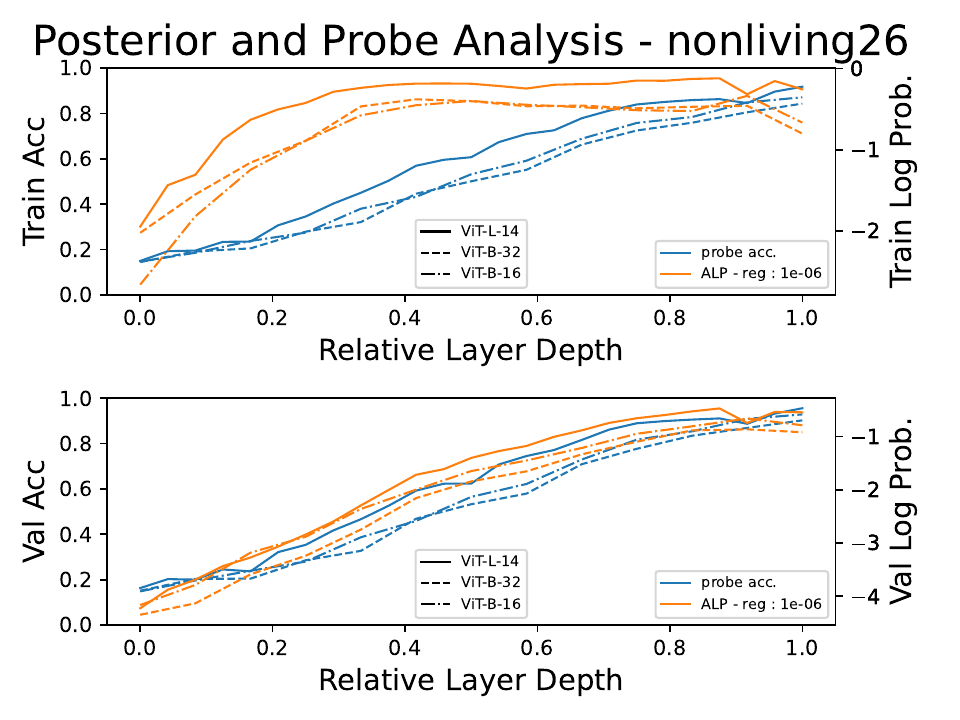}
        \caption{Regularization $10^{-6}$.}
        \label{fig:CLIPposteriorAnalysisBreeds-allnetperregnl26_1e-6_demo}
    \end{subfigure}
    \begin{subfigure}{0.48\textwidth}
        \centering
        \includegraphics[width=\linewidth]{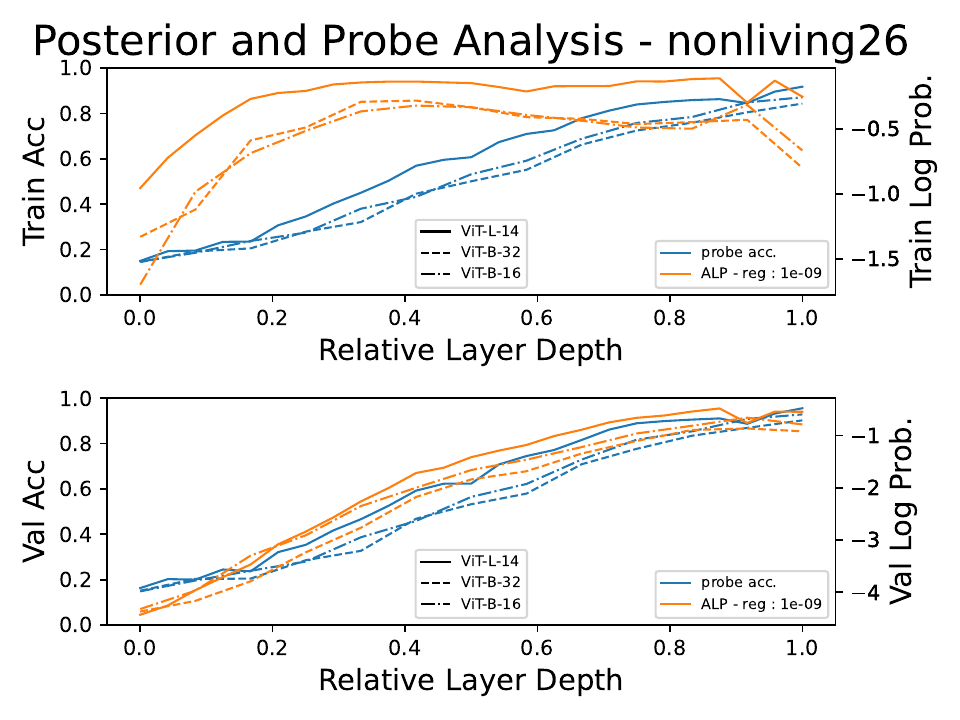}
        \caption{Regularization $10^{-9}$.}
        \label{fig:CLIPposteriorAnalysisBreeds-allnetperregnl26_1e-9_demo}
    \end{subfigure}
    \caption{Regularization-wise analysis of layerwise ALP vs. linear probe performance for nonliving 26 breeds dataset.}
    \label{fig:CLIPAnalysisBreeds-nonliving26-Part2}
\end{figure}

In this section, we break the constituents of \autoref{fig:entity13-allnetallreg}, \autoref{fig:CLIPAnalysisLivingNonLiving},  \autoref{tab:breedsCLIPpearcorr1} and \autoref{tab:breedsCLIPpearcorr2} into individual plots comparing the behavior of ALP and linear probe accuracy for each CLIP model on a given dataset for varying regularization schemes. A complete per regularization breakdown for the 3 CLIP Models  corresponding to entity-13 from \autoref{fig:entity13-allnetallreg} is provided in \autoref{fig:CLIPAnalysisBreeds-entity13-Part1} and \autoref{fig:CLIPAnalysisBreeds-entity13-Part2}, for entity-30 in \autoref{fig:entity30-allnetallreg}, the same is shown in  \autoref{fig:CLIPAnalysisBreeds-entity30-Part1} and \autoref{fig:CLIPAnalysisBreeds-entity30-Part2}. For living-17 and non-living-26 from \autoref{fig:living17-allnetallreg} and \autoref{fig:nonliving26-allnetallreg}, the analysis is shown in \autoref{fig:CLIPAnalysisBreeds-living17-Part1}, \autoref{fig:CLIPAnalysisBreeds-living17-Part2} and \autoref{fig:CLIPAnalysisBreeds-nonliving26-Part1}, \autoref{fig:CLIPAnalysisBreeds-nonliving26-Part2} respectively.

\end{document}